%% file: iclr2025_conference.tex
\PassOptionsToPackage{table}{xcolor}
\documentclass{article} % For LaTeX2e
\usepackage{iclr2025_conference,times}

\input{math_commands.tex}

\usepackage[table]{xcolor}
\usepackage{hyperref}    % Hyperlinks
\usepackage{url}         % URL handling
\usepackage{algorithm}    % For algorithms
\usepackage{algpseudocode} % For pseudocode
\usepackage{subcaption}   % For subfigures
\usepackage{booktabs}     % For formal tables
\usepackage{array}        % For tables
\usepackage{tabularx}     % For tables with adjustable column widths

\usepackage{amsthm}       % For theorems and definitions
\usepackage{graphicx}     % For including graphics
\usepackage{wrapfig}      % For wrapping figures
\usepackage{bbm}          % For bold math symbols
\usepackage{cleveref}     % For smart referencing
\usepackage{authblk}      % For multi-author affiliations

\definecolor{lightyellow}{rgb}{1, 1, 0.88}
\definecolor{lightgray}{rgb}{0.9, 0.9, 0.9}

\theoremstyle{definition}
\newtheorem{definition}{Definition}[section]
\newtheorem{assumption}{Assumption}[section]

\title{GAQAT: Gradient-adaptive Quantization-aware Training for Domain Generalization}

\author{
    \centering
    Jiacheng Jiang\textsuperscript{1},
    Yuan Meng\textsuperscript{1},
    Chen Tang\textsuperscript{1,2},
    Han Yu\textsuperscript{1},
    Qun Li\textsuperscript{1},
    Zhi Wang\textsuperscript{1},
    Wenwu Zhu\textsuperscript{1} \\
    \textsuperscript{1}Tsinghua University\\
    \textsuperscript{2}MMLab, CUHK \\
}

\iclrfinalcopy

\begin{document}

\maketitle

\input{./abstract}
\input{./intro}
\input{./preliminary}

\input{./method}

\input{./experiment}
\input{./related_work}
\input{./CONCLUSION_AND_FUTURE_WORK.tex}

\bibliography{iclr2025_conference}
\bibliographystyle{iclr2025_conference}

\end{document}

%% file: math_commands.tex
\usepackage{amsmath,amsfonts,bm}

\def\eqref#1{equation~\ref{#1}}
\def\1{\bm{1}}

\DeclareMathAlphabet{\mathsfit}{\encodingdefault}{\sfdefault}{m}{sl}
\SetMathAlphabet{\mathsfit}{bold}{\encodingdefault}{\sfdefault}{bx}{n}

%% file: abstract.tex
\begin{abstract}

Research on loss surface geometry, such as Sharpness-Aware Minimization (SAM), shows that flatter minima improve generalization. Recent studies further reveal that flatter minima can also reduce the domain generalization (DG) gap. However, existing flatness-based DG techniques predominantly operate within a full-precision training process, which is impractical for deployment on resource-constrained edge devices that typically rely on lower bit-width representations (e.g., 4 bits, 3 bits). Consequently, low-precision quantization-aware training is critical for optimizing these techniques in real-world applications.
In this paper, we observe a significant degradation in performance when applying state-of-the-art DG-SAM methods to quantized models, suggesting that current approaches fail to preserve generalizability during the low-precision training process. To address this limitation, we propose a novel Gradient-Adaptive Quantization-Aware Training (GAQAT) framework for DG. 
Our approach begins by identifying the scale-gradient conflict problem in low-precision quantization, where the task loss and smoothness loss induce conflicting gradients for the scaling factors of quantizers, with certain layers exhibiting opposing gradient directions. This conflict renders the optimization of quantized weights highly unstable. To mitigate this, we further introduce a mechanism to quantify gradient inconsistencies and selectively freeze the gradients of scaling factors, thereby stabilizing the training process and enhancing out-of-domain generalization.
Extensive experiments validate the effectiveness of the proposed GAQAT framework. On PACS, \textcolor{black}{both 3-bit and 4-bit exceed directly integrating DG and QAT by up to 4.5\%. On DomainNet, our 4-bit results deliver nearly lossless performance compared to the full-precision model, while achieving improvements of up to 1.39\% and 1.06\% over the SOTA QAT baseline for 4-bit and 3-bit quantized models, respectively.}

% “随着智能设备在医疗、自动驾驶等关键行业的广泛应用，数据的不确定性带来了潜在的重大风险，因此，模型在不同领域中的泛化能力变得至关重要。然而，目前大多数研究仍聚焦于全精度模型的泛化能力，而对于在边缘设备中广泛使用的量化模型的泛化能力研究相对较少。我们观察到，量化后的模型在域外场景中的性能下降更加明显。DG领域近期先进的SAM相关工作表明，平坦的损失曲面能够提升全精度模型的域外泛化能力，这启发我们思考是否可以通过优化量化模型的损失曲面平坦性来提高其域外泛化能力。

% 本文提出了一种适用于领域泛化场景的“选择性缩放因子梯度冻结量化感知训练框架”。具体而言，受到SAGM的启发，我们首次在领域泛化场景中引入了平坦性目标。然而，由于量化模型中的实际运算权重会受到缩放因子的影响，引入的平坦性目标与量化过程中缩放因子的梯度可能会产生冲突。为解决这一问题，我们通过选择性冻结缩放因子的梯度，确保训练过程的稳定性，从而找到更优的域外泛化解法。

% 广泛的实验结果表明，所提出的方法能够有效找到既适合量化又具备优秀域外泛化能力的参数。实验数据展示了该方法在关键性能指标上显著提升，例如，在某指标上取得了xx%的增长。”
\end{abstract}

%% file: intro.tex
\section{Introduction} 

Deep learning models have demonstrated remarkable performance across various computer vision tasks, such as classification~\citep{he2016deep,sandler2018mobilenetv2,dosovitskiy2020image}, detection~\citep{zhu2020deformable,zhang2022dino}, and semantic segmentation~\citep{zhou2022rethinking,strudel2021segmenter}. 
However, these models typically experience significant performance degradation in real-world applications due to domain shift, which manifests as poor generalization to previously unseen data distributions. 
Domain generalization (DG) seeks to address this challenge by enabling models trained on observed source domains to generalize effectively to unseen target domains. 
Strategies such as domain alignment \citep{li2018domain,muandet2013domain}, data augmentation \citep{zhou2021domain,volpi2018generalizing}, and meta learning \citep{li2018learning,balaji2018metareg} are commonly employed techniques. 
Recent studies \citep{gulrajani2020search}, however, indicate that despite the development of these sophisticated techniques, basic empirical risk minimization (ERM) still yields comparable out-of-distribution generalization when experimental conditions are carefully controlled. 
Concurrently, growing attention has been directed towards the geometry of the loss landscape~\citep{li2024enhancing,foret2020sharpness,andriushchenko2022towards,wen2023sharpness} in generation, particularly the Shareness-aware Minimization (SAM) that pursues flatter minima during training. Recent works~\citep{cha2021swad, wen2023sharpness} has shown that a flatter minimum could lead to a smaller DG gap. Inspired by previous studies of flat minima~\citep{izmailov2018averaging,foret2020sharpness,liu2022towards,zhuang2022surrogate,zhang2023gradient,wang2023sharpness}, flatness-aware methods start to gain attention and exhibit remarkable performance in domain generalization. 

Despite the demonstrated effectiveness of flatness-aware methods in improving out-of-domain generalization, they are confined to \emph{full-precision training}, which means the resulting models of current methods are not very practical to deploy. 
In other words, in many real-world applications, especially those involving deployment on edge devices and are truly vulnerable to domain shift environments, models operate under very computationally-constrained resources. 
Although the trained low-precision computations, a.k.a. the quantization-aware training~\citep{zhou2016dorefa,tang2022mixed,esser2019learned}, have been extensively studied in I.I.D research for improving the runtime efficiency, in which the models are trained with simulated quantization during the forward-backward process and thus the weights can be aware of the numerical change, there still are challenging to achieve the generalized quantization-aware training for domain generalization, as 
\emph{(a) distinct objectives:} Low precision aims to reduce model complexity, but conflicts with maintaining generalization.
% how to properly formulate the optimization problem for the models that achieve two distinct learning objectives, 
and 
\emph{(b) training instability}: how to ensure the proper convergence for the low-precision weights as the simulated quantization and sharpness-aware minimization both involve specific gradient approximation~\citep{wen2023sharpness,nagel2022overcoming,tang2024retraining}. 
In fact, we have observed when directly applying DG-SAM methods~\citep{wen2023sharpness} to quantization-aware training~\citep{esser2019learned,zhou2016dorefa}, there could be an unexpected degradation of the model's generalization performance (e.g., the average out-of-domain performance drops by 28.36\% when quantized to 4 bits in PACS).
% This counterintuitive outcome suggests that naïve application neglects the critical interaction between quantization and generalization. 

\begin{figure}[t]
    \centering
    \includegraphics[width=1\linewidth]{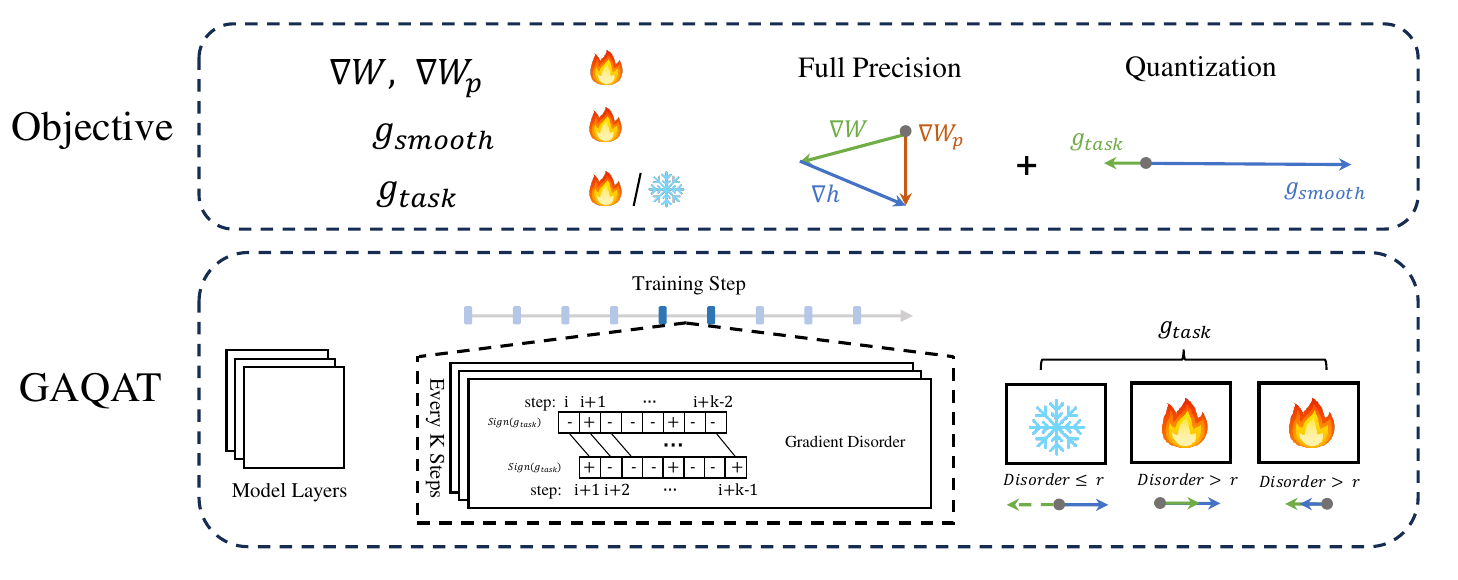}
    \caption{Illustration of GAQAT. 
    % We introduce a smoothing objective during quantization, generating new scale gradients. 
    Compared to full-precision weight gradients, the tensor-wise scale gradients have only two directions: positive and negative. For the newly introduced task-related scale gradients, we apply the GAQAT method for selective freezing. We calculate the disorder of each scale's task gradient \( \mathbf{g}_{{\text{task}}} \) and freeze those with disorder below a certain threshold to improve the model's generalization ability.}
    \label{fig:process} 
\end{figure}

In this paper, we propose the Gradient-Adaptive Quantization-Aware Training (GAQAT) framework for domain generalization. 
Specifically, we first incorporate the smoothing 
factor term into the quantizer to ensure that both quantization and smoothness can be optimized jointly. 
Though the optimization objective seems reasonable and is optimizable, the quantizer receives two distinct gradients of the quantization and sharpness-aware minimization. 
By conducting a thorough analysis of the behavior of the quantizer gradients, we reveal that the significant conflicts between task loss (empirical loss) and smoothness loss induced by the gradient approximations cause the generalization ability of the trained model to degrade, even worse-performing than models optimizing a single objective. 
To this end, we define the \emph{gradient disorder} that depicts the inconsistency of gradient directions during
training to quantify the magnitudes of gradient conflicts. 
Based on this, we further design a dynamic freezing strategy, which selectively enables or disables the update of quantizers according to their gradient disorders, thus ensuring global convergence for the overall performance. 
The illustration of the proposed method is shown in Figure~\ref{fig:process}. 
% We conduct extensive experiments on two representative benchmarks, PACS~\citep{Li_2017_ICCV} and DomainNet~ \citep{peng2019domain}, to demonstrate the effectiveness of GAQAT. 

In summary, we have made the following contributions: 
\begin{itemize}
    \item We propose GAQAT, a framework to achieve efficient domain generalization by considering low-precision computations. 
    For the first time we can empower the quantized model with good out-of-distribution generalization. 
    % By a thorough analysis, we further identify the optimization process exists significant gradient conflicts between the task loss and smoothness loss, which degrade generalization. 
    \item We introduce the concept of gradient disorder to quantify gradient conflict magnitudes during optimization. 
    Building on this, we design a dynamic freezing strategy that selectively updates quantizers based on gradient disorder, ensuring global convergence and improved generalization performance. 
    \item Extensive experiments on PACS and DomainNet demonstrate the effectiveness of GAQAT. Specifically, on PACS, 4-bit accuracy reaches 61.33\%, surpassing the baseline by 4.4\%. In 3-bit, it still exceeds the baseline by 4.55\%. On DomainNet, 4-bit achieves 40.74\%, close to the full precision accuracy of 40.95\%, while 3-bit reaches 39.53\%, still outperforming the baseline.
\end{itemize}

%% file: preliminary.tex
\section{Preliminaries}
\label{gen_inst}

\subsection{Quantization}

% 加上iid的元素，解释下
% Uniform quantization is favored in neural networks for its hardware efficiency. 
% The range of full-precision value $x$ is divided into equal intervals, and each interval is assigned a quantization level.
% Each layer employs two distinct scaling factors to accommodate varying distributions of weights and activations, facilitating efficient inference. This approach converts full-precision activations and weights into low-precision integers.
% In \(b\)-bit quantization, activations are limited to the range \([0, 2^b - 1]\) and weights to \([-2^{b-1}, 2^{b-1} - 1]\). 

We consider the uniform quantization function for both weight and activation of layers: 
\(
\hat{\mathbf{v}} = Q_b(\mathbf{v}; s) = s \times \left\lfloor \text{clip}\left( \frac{\mathbf{v}}{s}, l, u \right) \right\rceil,
\)
where \(\lfloor \cdot \rceil\) denotes round-to-nearest operator, \(s\) is a learnable scaling factor in QAT~\citep{esser2019learned,tang2022mixed}, and the \(\texttt{clip}\) function ensures values stay within the bounds \([l, u]\). 
In \(b\)-bit quantization, for activation quantization, we set \(l = 0\) and \(u = 2^b - 1\); for weight quantization, we set \(l = -2^{b-1}\) and \(u = 2^{b-1} - 1\). 
Furthermore, to overcome the non-differentiability of the rounding operation, the Straight-Through Estimator (STE)~\citep{bengio2013estimating} is employed to approximate the gradients: \(
\frac{\partial \mathcal{L}}{\partial \mathbf{v}} \approx \frac{\partial \mathcal{L}}{\partial \hat{\mathbf{v}} } \cdot 1_{l \leq \frac{\mathbf{v}}{s} \leq u},
\). 

% Due to the non-differentiability of the rounding operation, the Straight-Through Estimator (STE) \yym{add cite} is employed to approximate the gradients.
% % , ensuring smooth training. 
% The gradient approximation is given by:
% \(
% \frac{\partial L}{\partial x} \approx \frac{\partial L}{\partial \hat{x}} \cdot 1_{l \leq \frac{x}{s} \leq u},
% \)
% where \(1_A\) is an indicator function that is 1 if the element is within set \(A\) and 0 otherwise. 

\subsection{Flatter Minima in Domain Generalization}

Following SAGM~\citep{wang2023sharpness}, we adopt three objectives for sharpness-aware minimization over the observed domains $D$: 
(a) empirical risk \( \mathcal{L}_{ER}(\theta; D) \), 
(b) perturbed loss \( \mathcal{L}_p(\theta; D) \), and 
(c) the surrogate gap \( h(\theta) \):= \( \mathcal{L}_{p}(\theta; D) - \mathcal{L}_{ER}(\theta; D) \). 
Minimizing \( \mathcal{L}_{ER}(\theta; D) \) and \( \mathcal{L}_p(\theta; D) \) finds low-loss regions, while minimizing \( h(\theta) \) ensures a flat minimum. This combination improves both training performance and generalization. 
Hence, the overall optimization is:\(\min [\mathcal{L}_{ER}(\theta; D) + \mathcal{L}_p(\theta - \alpha \nabla \mathcal{L}_{ER}(\theta; D); D)] \)
where \( \alpha \) is the hyperparameter, which can be rewritten as:
\(
\min \mathcal{L}(\theta; D) + \mathcal{L}(\theta + \hat{\epsilon} - \alpha \nabla \mathcal{L}(\theta; D); D)
\)
with \( \hat{\epsilon} = \rho \frac{\nabla \mathcal{L}(\theta; D)}{\|\nabla \mathcal{L}(\theta; D)\|} \).

%% file: method.tex
\section{Method}
\label{sec:method}

\subsection{Quantization in DOMAIN GENERALIZATION}
% 解释dg处理的问题，解释最近很出名的是sam相关的，但全精度用不了边缘设备，所以需要ood量化, 给出我们的公式，然后解释sam的梯度运算，说明我们新引入了scale梯度
% 给出我们的公式  然后解释梯度的来源然后解释

% 由于此前没有工作探索过DG场景下的量化流程，所以我们在这里定义我们的流程，如图x, 我们会先有一个已经预训练好的，有一定泛化性能的模型\theta_0, 然后在QAT的过程中，我们只能使用the source domain data \( D_s \)，可以有一个或者多个，然后受到sam及其变体的启发，我们也对量化目标引入了平滑因子，以求获得更好的泛化性能，最后我们通过源数据和现有的优化目标进行QAT训练，获得泛化性能良好的量化模型，其中泛化性在目标域\( D_d \)上进行测试。the loss function is defined as \(\min L(Q(\theta); D) + L_p(Q(\theta - \alpha \nabla L(Q(\theta); D); D))\)
% Given the lack of prior research on quantization processes in Domain Generalization (DG) scenarios, we establish our approach, illustrated in Figure ~\ref{fig:process}. We start with a pre-trained model, \( \theta_0 \), which possesses initial generalization capabilities. 
% During QAT, we exclusively utilize source domain data \( D_s \), which may encompass one or more datasets. SAM and its variants, 

Firstly, we incorporate the smoothing factor into the quantizer to perform the generalization optimization within the latent weight space. 
Then, we directly employ quantization-aware training with source domains. The loss function is defined as: 
\begin{equation}
    \min \mathcal{L}_{ER} \left( Q \left(\theta; \mathbf{s}_w\right); D \right)  + \mathcal{L}_p \left( Q \left (\theta - \alpha \nabla \mathcal{L} \left(Q 
 \left(\theta; \mathbf{s}_w \right); D \right); \mathbf{s}_w \right); D \right) 
    \label{eq:sagm_qat} 
\end{equation}
However, we have observed that directly adopting this objective can lead to performance degradation, as shown in Table~\ref{tab:pacs} and Table~\ref{tab:domainnet}.

\subsection{Analysis of the Quantizer Gradient Conflict Issue}
% 引入平滑因子之后带来性能掉点让人困惑，这是低精度所没有出现的现象，我们认为必然和量化的特性因子有关，为了具体解释\( \mathbf{g}_{{\text{smooth}}} \)和\( \mathbf{g}_{{\text{task}}} \) 之间的干扰，我们可视化了训练过程中的两个梯度和，如图x, 可以看到显著的梯度冲突现象。
% The performance drop after introducing the smoothing factor is puzzling, as this phenomenon does not occur in the low-precision setting. 
Compared to full-precision training, Eq.~(\ref{eq:sagm_qat}) has several scale factors $s_{*}$ in the quantizers that will correspond to two optimization targets, thus producing two sets of gradients. 
One set is the original task-related gradient, which we abbreviate as \( \mathbf{g}_{{\text{task}}} \) from $\mathcal{L}_{ER} (\cdot)$, and the other is the newly introduced flatness-related gradient, abbreviated as \( \mathbf{g}_{{\text{smooth}}} \) from $\mathcal{L}_{p} (\cdot)$. 
% We believe it is inherently related to the quantization-specific factors. 

\begin{figure}[t]
    \centering
    \includegraphics[width=\linewidth]{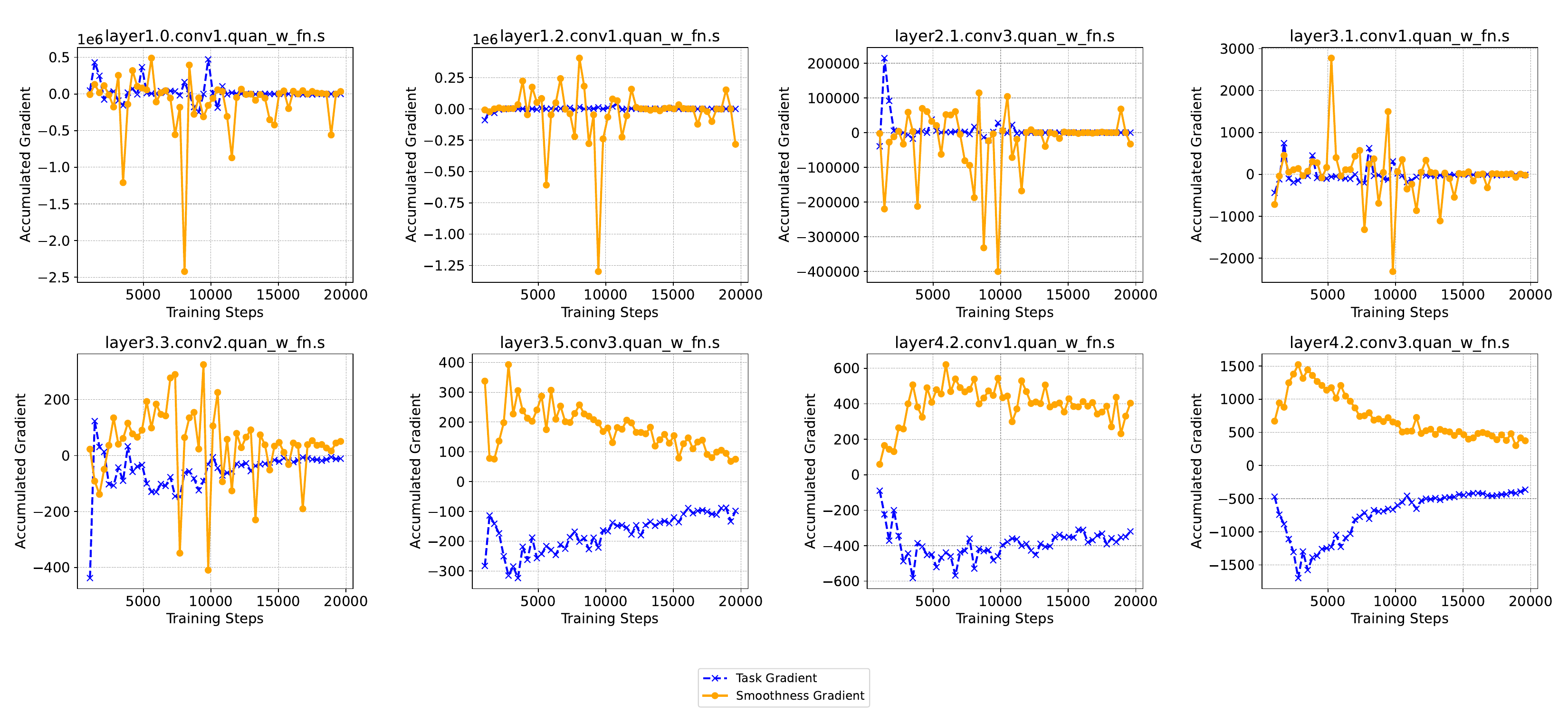}
    \caption{Results of cumulative gradients every 350 steps in the 4-bit test on the PACS ART domain, revealing conflicts in the scaling factors.}
    \label{fig:conflict}
\end{figure}

However, the scale factor, used to portray the characteristic of weight and activation distribution \citep{tang2022mixed}, is highly sensitive to the perturbations \citep{esser2019learned,liu2023oscillation}. 
We therefore have the following hypothesis for the scaling factor in quantizer: 
% (Incomplete Scaling Factor Convergence) 
\emph{
The apparent convergence of scaling factors reaching a sub-optimal state does not necessarily indicate satisfactory convergence and can negatively impact OOD performance.
}
To verify this hypothesis, we perform perturbations on the scales of certain layers in the trained model by further scaling them by $x \in \{0.8, 0.9, 1.1, 1.2\}$ times. 
As shown in Table~\ref{tab:performance_comparison}, perturbing the scale to certain layers significantly improves OOD performance, while in other layers, it results in performance degradation. 
This indicates the proper convergence of quantization parameters (the scaling factor in the quantizer) is of importance for out-of-distribution generalization, proving that the scale converges suboptimally due to the conflicted gradients of two objectives. 
\begin{table}[t]
\centering
\caption{Performance results for perturbed scaling factors in the 4-bit test on Clipart and Infograph datasets from DomainNet. The notation x\% indicates a scaling factor change by x\%. \textcolor{black}{Red highlights performance degradation, while green signifies improvement.} These results suggest that the apparent convergence of scaling factors towards a suboptimal state does not necessarily imply satisfactory convergence and can negatively affect OOD performance.}
\label{tab:performance_comparison}
\small % 将表格尺寸缩小
\begin{tabular}{lccccc}
\hline
\textbf{Layer} & \textbf{origin} & \textbf{80\%} & \textbf{90\%} & \textbf{110\%} & \textbf{120\%} \\
\hline
layer3.0.conv1.w.s & 60.21 / 15.81 & \textcolor{green}{60.30} / \textcolor{green}{15.93} & \textcolor{red}{60.15} / \textcolor{green}{15.94} & \textcolor{red}{59.96} / \textcolor{red}{15.62} & \textcolor{red}{59.82} / \textcolor{red}{15.38} \\
layer3.0.conv1.a.s & 60.21 / 15.81 & \textcolor{green}{60.47} / \textcolor{green}{16.12} & \textcolor{green}{60.31} / \textcolor{green}{15.90} & \textcolor{red}{60.10} / \textcolor{red}{15.72} & \textcolor{red}{59.93} / \textcolor{red}{15.65} \\
layer1.0.conv1.w.s & 60.21 / 15.81 & \textcolor{green}{60.25} / \textcolor{red}{15.60} & \textcolor{red}{60.14} / \textcolor{red}{15.61} & \textcolor{green}{60.32} / \textcolor{red}{15.48} & \textcolor{red}{60.18} / \textcolor{red}{15.27} \\
layer1.0.conv1.a.s & 60.21 / 15.81 & \textcolor{green}{60.23} / 15.81 & \textcolor{green}{60.22} / \textcolor{green}{15.85} & \textcolor{green}{60.26} / \textcolor{red}{15.78} & \textcolor{green}{60.24} / \textcolor{red}{15.67} \\
\hline
\end{tabular}
\end{table}
To further show the interference between \( \mathbf{g}_{{\text{smooth}}} \) and \( \mathbf{g}_{{\text{task}}} \), we visualized the sum of these two gradients during the training process. 
As shown at the top of Figure~\ref{fig:conflict}, a significant gradient conflict is evident. 
Morever, for certain layers, \( \mathbf{g}_{{\text{task}}} \) and \( \mathbf{g}_{{\text{smooth}}} \) is opposite and tend to cancel each other out~(bottom of Figure~\ref{fig:conflict}). \textcolor{black}{This suggests that the scaling factors of these layers are approaching a state we define as the sub-optimal equilibrium state.}
Since both simulated quantization and sharpness-aware minimization involve specific gradient approximations and according to \citep{liu2023oscillation}, the weight oscillations caused by the discrete nature of quantization can be significantly amplified by  learnable scaling factors, the conflict between \( \mathbf{g}_{{\text{task}}} \) and \( \mathbf{g}_{{\text{smooth}}} \) can substantially negatively impact the performance of QAT in DG scenarios.

% \begin{figure}[t]
%     \centering
%     \begin{minipage}[b]{\linewidth}
%         \centering
%         \includegraphics[width=\linewidth]{conflict.pdf}
%         \caption{Cumulative gradients of scaling factors showing conflict between \( \mathbf{g}_{{\text{task}}} \) and \( \mathbf{g}_{{\text{smooth}}} \).}
%         \label{fig:conflict}
%     \end{minipage}
    
%     \vfill % Adds vertical space between the two images
    
%     \begin{minipage}[b]{\linewidth}
%         \centering
%         \includegraphics[width=\linewidth]{opposite.pdf}
%         \caption{Opposing gradients between \( \mathbf{g}_{{\text{task}}} \) and \( \mathbf{g}_{{\text{smooth}}} \) in certain layers.}
%         \label{fig:opposite}
%     \end{minipage}
% \end{figure}

\begin{figure}[t]
    \centering
    
    \includegraphics[width=\linewidth]{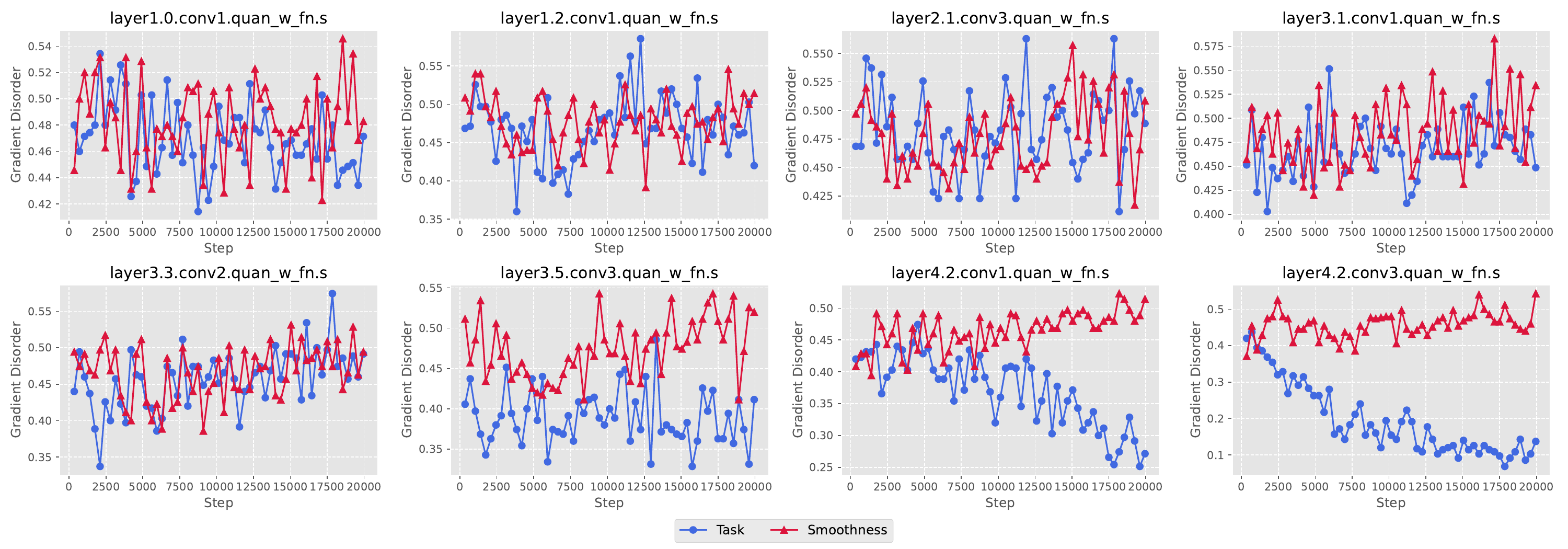}
    \caption{Results of task and smoothness gradient disorder of scaling factors over 350 steps in the 4-bit test on the PACS ART domain, revealing in some layers, the gradient disorder of the \(\mathbf{g}_{{\text{task}}}\) decreases significantly as training progresses.}
    \label{fig:disorder}
\end{figure}

% As depicted in Figure~\ref{fig:opposite}, during training, we notice that in certain layers, the task-related and flatness-related scaling factor gradients are opposite and tend to balance each other. This suggests that these layers' scaling factors are reaching a state of equilibrium, which is 

\subsection{Selective Freezing to Resolve Gradient Conflicts}

To address the issue of scaling factor gradient conflicts, we propose Gradient-Adaptive Quantization-Aware Training (GAQAT) framework for domain generalization, a selective freezing training strategy. First, we define the \textit{gradient disorder} to quantify the inconsistency of gradient directions during training.

\begin{definition}
\textbf{Gradient Disorder:} Suppose we have \(K\) steps of training, and at each step \(j\), this step's gradient is formalized as \(\mathbf{g}_j\). We define two gradient sequences:\textcolor{black}{ \(S_1 = \{\mathbf{g}_1, \mathbf{g}_2, \dots, \mathbf{g}_{K-1}\}\) and \(S_2 = \{\mathbf{g}_2, \mathbf{g}_3, \dots, \mathbf{g}_K\}\)}. Let \(\operatorname{sgn}(\cdot)\) denote the element-wise sign function. The gradient disorder is defined as:
\textcolor{black}{\begin{equation}
% \large
\delta = \frac{1}{K}  \mathbbm{1} \left( {\mathbf{\operatorname{sgn}(S_1) \neq \operatorname{sgn}(S_2)}} \right),  
% \operatorname{num}(\operatorname{sgn}(S_1) \neq \operatorname{sgn}(S_2))
\end{equation}}
where $\mathbbm{1}(\cdot)$ is the indicator function. 
$\delta$ represents the proportion of steps where the gradient direction is opposite to that of the previous step.
\end{definition}
% \begin{figure}[htbp]
%     \centering
%     \begin{minipage}[b]{0.45\linewidth}
%         \includegraphics[width=\linewidth]{freeze1.png}
%         \caption{Gradient of a scaling factor with and without freezing the accuracy gradient.}
%         \label{fig:freeze1}
%     \end{minipage}
%     \hfill % Add some horizontal spacing
%     \begin{minipage}[b]{0.45\linewidth}
%         \includegraphics[width=\linewidth]{freeze2.png}
%         \caption{Gradient fluctuations of other scaling factors after freezing the accuracy gradient.}
%         \label{fig:freeze2}
%     \end{minipage}
% \end{figure}
\begin{wrapfigure}{r}{0.5\textwidth} % 将minipage与wrapfigure结合
    \begin{minipage}{0.5\textwidth} % 算法宽度50%
        \begin{algorithm}[H]
        \caption{Dynamic Selective Freezing Strategy for Scaling Factors}
        \label{alg:freeze}
        \begin{algorithmic}[1]
        \scriptsize
        \Require Training steps \(T\), evaluation interval \(K\), disorder threshold \(r\), set of scaling factors \(\{S_1, S_2, \dots, S_n\}\)
        \State Initialize step counter \(t \gets 0\), \(\text{freeze}[S_i] \gets \text{False}\) for all \(S_i\)
        \While{\(t < T\)}
            \For{each scaling factor \(S_i\)}
                \If{\(\text{freeze}[S_i] = \text{True}\)}
                    \State Update \(S_i\) using only \( \mathbf{g}_{{\text{smooth}}} \)
                \Else
                    \State Update \(S_i\) using both \( \mathbf{g}_{{\text{task}}} \) and \( \mathbf{g}_{{\text{smooth}}} \)
                \EndIf
            \EndFor
            \If{\(t \bmod K = 0\)}
                \For{each scaling factor \(S_i\)}
                    \State Compute gradient disorder \(\delta_{t,S_i}\)
                    \If{\(\delta_{t,S_i} < r\)}
                        \State \(\text{freeze}[S_i] \gets \text{True}\)
                    \Else
                        \State \(\text{freeze}[S_i] \gets \text{False}\)
                    \EndIf
                \EndFor
            \EndIf
            \State \(t \gets t + 1\)
        \EndWhile
        \end{algorithmic}
        \end{algorithm}
    \end{minipage}
\end{wrapfigure}

A lower gradient disorder indicates more consistent gradient directions, which implies more stable training. It is important to note that while a high disorder does not necessarily indicate incorrect gradients, a low disorder can provide some assurance of gradient correctness.

\textcolor{black}{Figure~\ref{fig:disorder} }indicates that in some layers, the gradient disorder of the \(\mathbf{g}_{{\text{task}}}\) decreases significantly as training progresses. This suggests that the gradient direction of the \(\mathbf{g}_{{\text{task}}}\) becomes increasingly consistent, which is somewhat counterintuitive. In contrast, the gradient disorder of the flatness scaling factor shows no significant change across layers. \textcolor{black}{And layers with lower task gradient disorder (as shown in the three images at the bottom-right in Figure~\ref{fig:disorder}) exhibit a clear phenomenon of opposite and similar-magnitude gradients in Figure~\ref{fig:conflict}. This indicates that layers with lower task gradient disorder are more likely to settle into sub-optimal equilibrium state.}

These observations suggest that the training of the \(\mathbf{g}_{{\text{task}}}\) gradients may interfere with the training of the flatness scaling factor. Inspired by the gradient freezing strategies \citep{liu2023oscillation,tang2024retraining,nagel2022overcoming}, we propose discarding \(\mathbf{g}_{{\text{task}}}\) in certain scales to mitigate these conflicts.

\begin{assumption}
\textbf{Impact of Incomplete Scaling Factor Convergence to other layers:} The apparent convergence of scaling factors reaching a suboptimal equilibrium state between task and flatness objectives could impact other layers, including causing outlier gradients.
\end{assumption}

\begin{figure}[t]
    \centering
    \includegraphics[width=\linewidth]{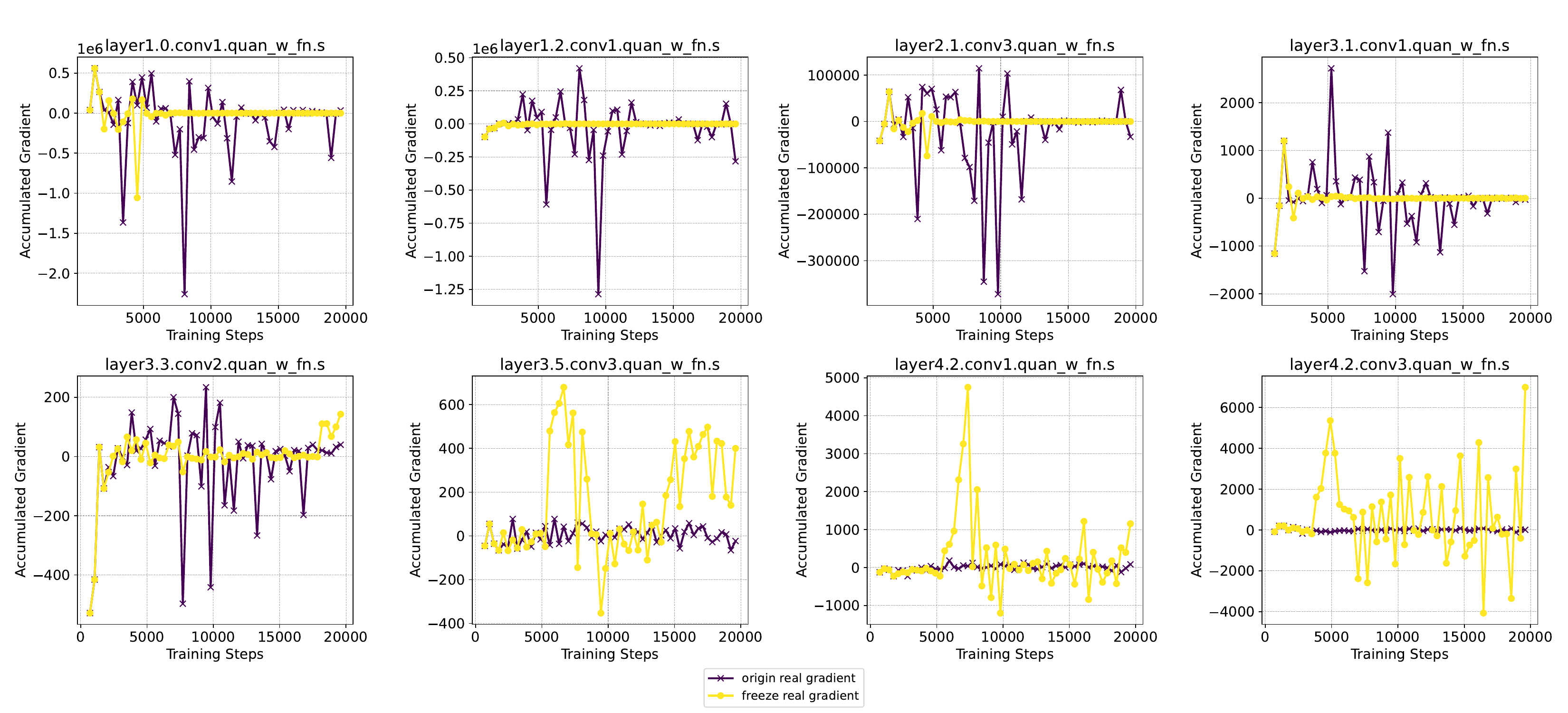}
    \caption{Results of freezing over 350 steps in the 4-bit test on the PACS ART domain, using gradient disorder as an indicator, with no unfreezing. The findings suggest that instability in gradient fluctuations is partly caused by interference between scaling factors during training. Moreover, the gradient disorder indicator proves to be a useful metric for determining when to freeze.}
    \label{fig:vs}
\end{figure}

To verify this hypothesis, we conducted an experiment using the gradient disorder of \(\mathbf{g}_{{\text{task}}}\) as an indicator of convergence (see \textcolor{black}{Figure~\ref{fig:vs}}). %Figure~\ref{fig:freeze1} shows the gradient changes of a scaling factor under frozen and unfrozen conditions, while Figure~\ref{fig:freeze2} displays the gradient fluctuations of other scaling factors after freezing the \(\mathbf{g}_{{\text{task}}}\).
The results demonstrate that the frozen scaling factor continues to be updated via \(\mathbf{g}_{{\text{smooth}}}\), and the gradient fluctuations in unfrozen layers are significantly reduced. This suggests that the instability in gradient fluctuations is partly caused by interference between scaling factors during training.

Based on these findings, we propose a selective freezing strategy to address scaling factor instability and improve flatness convergence. \textcolor{black}{Persistently freezing the \(\mathbf{g}_{{\text{task}}}\) of certain layers without selectively unfreezing them in specific cases may result in suboptimal  convergence.} Therefore, we adopt a dynamic approach. Every \(K\) steps, we evaluate the gradient disorder. If the disorder \(\delta_{t,S_i}\) for scaling factor \(S_i\) at step \(t\) is below a threshold \(r\), we freeze the \(\mathbf{g}_{{\text{task}}}\) of \(S_i\) for the next \(K\) steps; otherwise, we continue updating it. This dynamic selective freezing strategy allows the flatness of scaling factor to continue training while mitigating the adverse effects of gradient conflicts. By periodically reassessing and adjusting which scaling factors are frozen, we aim to improve overall convergence and enhance the model's generalization performance in DG scenarios. Full process is summarized in Algorithm~\ref{alg:freeze}.

%% file: experiment.tex
\section{Experiment}
\label{sec:experiment}
\subsection{Experimental Setup and Implementation Details}
\textbf{Quantization.} We follow established practices in Quantization-Aware Training (QAT) literature by employing the LSQ-type method \citep{esser2019learned} to quantize both weights and activations. The quantization scaling factors are learned with a fixed learning rate of \(1 \times 10^{-5}\). We use Mean Squared Error (MSE) range estimation \citep{nagel2021whitepaperneuralnetwork} to determine the quantization parameters for weights and activations. \textcolor{black}{Due to the risk of test data information leakage of supervised pretrained weights revealed by \citet{yu2024rethinking}, we employ MoCo-v2 \citep{chen2020improvedbaselinesmomentumcontrastive} pretrained ResNet-50 as initialization as recommended. Then we fine-tune the model using Empirical Risk Minimization (ERM) to obtain a full-precision model with generalization capabilities, which serves as the baseline for quantization.} \textcolor{black}{The weights and activations are fully quantized, except for the first convolutional layer, which quantizes only the activations, and the final linear layer, which remains unquantized,} striking a balance between efficiency and model capacity.
We evaluate the performance under extremely low bit-width conditions of 3 and 4 bits. 

\textbf{Datasets and evaluation protocol.} We conduct a comprehensive evaluation on two widely used DG datasets: PACS \citep{Li_2017_ICCV}, containing 9,991 images across 7 categories and 4 domains, and DomainNet \citep{peng2019domain}, consisting of 586,575 images across 345 categories and 6 domains. We basically follow the evaluation protocol of DomainBed \citep{gulrajani2020search}, including the optimizer, data split, and model selection, where we adopt test-domain validation as our model selection strategy for all algorithms in our experiments. 
For PACS, for each time we treat one domain as the test domain and other domains as training domains, which is the leave-one-domain-out protocol commonly adopted in DG.  
For DomainNet, following \citet{yu2024rethinking}, we divide the domains into three groups: (1) \textit{Clipart} and \textit{Infograph}, (2) \textit{Painting} and \textit{Quickdraw}, and (3) \textit{Real} and \textit{Sketch}. Then we employ the leave-one-group-out protocol, where we treat one group of two domains as test domains and other two groups as training domains each time. 
For the number of training steps, for full-precision models we set it as 5,000 for PACS and 15,000 for DomainNet following \citet{cha2021swad}, while for quantization training we use 20,000 for PACS and 50,000 for DomainNet. 
To reduce time cost, for quantization training we conduct validation and testing for DomainNet only after 45,000 steps. 

% We consistently select the model corresponding to the hyperparameters that achieve the best performance on the validation set for testing and choose the best performance on the test set to report.

\textbf{Hyperparameter settings.} 
Given the substantial computational resources required by the original DomainBed setup, we adjust the hyperparameter search space and conduct grid search to reduce computational cost following SAGM \citep{wang2023sharpness}. 
The search space of learning rate is \{1e-5, 3e-5, 5e-5\}, and the dropout rate is fixed as zero. 
The batch size of each training domain is set as $32$ for PACS and $24$ for DomainNet. 
Following SAM \citep{foret2020sharpness}, we fix the hyperparameter $\rho=0.05$. 
Following SAGM \citep{wang2023sharpness}, we set $\alpha$ in \Cref{eq:sagm_qat} as $0.001$ for PACS and $0.0005$ for DomainNet, and set weight decay as 1e-4 for PACS and 1e-6 for DomainNet. 

For PACS, the gradient disorder threshold \(r\) is selected from \(\{0.28, 0.30, 0.32\}\) for both 3-bit and 4-bit quantization. The number of freeze steps is selected from \(\{300, 350, 400\}\) for 4-bit quantization, and from \(\{100, 150, 200\}\) for 3-bit quantization. 
For DomainNet, \(r\) is selected from \(\{0.20, 0.25\}\) for 4-bit quantization, and from \(\{0.02, 0.03\}\) for 3-bit quantization. The number of freeze steps is chosen from \(\{3000, 4000\}\) for 4-bit quantization, and from \(\{200, 300\}\) for 3-bit quantization, as we observed that conflicts are more severe in 4-bit than in 3-bit quantization.
To reduce the high computational cost, we first select the shared hyperparameters, i.e. learning rate, weight decay, through grid search, which serve as the base hyperparameter configuration. Then we fix the base configuration and conduct further grid search on our specific hyperparameters, i.e. freeze steps, freeze threshold.

\subsection{Main Results}

We evaluated our method on the PACS and DomainNet datasets, comparing it to existing approaches (see Tables~\ref{tab:pacs} and \ref{tab:domainnet}). Our method achieves the best performance across different quantization bit-widths (4/4 and 3/3). At 4-bit quantization, it attains an average test accuracy of \textbf{61.33\%} on PACS, outperforming LSQ (\textbf{58.98\%}) and SAGM+LSQ (\textbf{56.93\%}); When the quantization bit-width is reduced to 3 bits, our method maintains superior performance with an average accuracy of \textbf{57.13\%}, remain the best, demonstrating its robustness.

\begin{table}[h!]
\centering
\caption{Results on PACS dataset.}
\label{tab:pacs}
\scriptsize
\renewcommand{\arraystretch}{1.1}
\resizebox{\textwidth}{!}{
\begin{tabular}{@{}lcccccc@{}}
\toprule
\rowcolor{lightgray}
Method & Bit-width (W/A) & Art (val/test) & Cartoon (val/test) & Photo (val/test) & Sketch (val/test) & Avg (val/test) \\
\midrule
ERM & Full & 96.63/84.62 & 95.79/80.86 & 96.78/95.73 & 96.48/79.96 & 96.42/85.29 \\
\midrule
LSQ & 4/4 & \textbf{88.28}/\textbf{51.07} & \textcolor{black}{\textbf{78.74}}/58.10 & 80.81/63.77 & 74.96/62.98 & 80.70/58.98 \\
SAGM+LSQ & 4/4 & 86.21/46.49 & 76.86/55.12 & 81.79/64.67 & 70.60/61.45&  78.87/56.93\\
\rowcolor{lightyellow}
Ours & 4/4 & 86.75/49.24& 78.11\textcolor{black}{/\textbf{59.22}}& \textbf{85.31}/\textbf{69.46} & \textbf{77.25}/\textbf{67.40} & \textbf{81.86}/\textbf{61.33}\\
\midrule
LSQ & 3/3 & 82.07/39.29 & 74.97/\textbf{58.69} & \textbf{79.21}/59.28 & 74.88/\textbf{64.41} & 77.78/55.42 \\
SAGM+LSQ & 3/3 & 83.48/43.56 & 72.34/52.45& 74.22/58.16 & 64.94/56.14& 73.75/52.58\\
\rowcolor{lightyellow}
Ours & 3/3 & \textbf{84.43}/\textbf{44.36} & \textbf{75.77}/59.06& 76.85/\textbf{61.75} & \textbf{75.70}/63.33& \textbf{78.19}/\textbf{57.13}\\
\bottomrule
\end{tabular}
}
\end{table}
\begin{table}[h!]
\centering
\caption{Results on DomainNet dataset.}
\label{tab:domainnet}
\tiny
\renewcommand{\arraystretch}{1.1}
\begin{tabularx}{\textwidth}{@{}lXccccccc@{}}
\toprule
\rowcolor{lightgray}
Method & Bit-width (W/A) & Clipart & Infograph & Painting & Quickdraw & Real & Sketch & Avg \\
\midrule
ERM & Full & 66.80/59.42 & 66.80/18.30 & 61.13/47.90 & 61.13/13.78 & 58.20/57.82 & 58.20/48.46 & 62.04/40.95 \\
\midrule
LSQ & 4/4 & 66.34/60.45 & 66.34/15.65 & 59.56/44.69 & 59.56/14.76 & 57.82/52.70 & 57.82/47.82 & 61.24/39.35  \\
SAGM+LSQ & 4/4 & 65.77/60.73 & 65.77/15.64 & 61.21/46.67 & 61.21/16.29 & 56.77/52.22 & 56.77/48.45 & 61.25/40.00  \\
\rowcolor{lightyellow}
Ours & 4/4 & \textbf{67.20}/\textbf{61.00} & \textbf{67.20}/\textbf{16.12} & \textbf{62.51}/\textbf{47.80} & \textbf{62.51}/\textbf{16.44} & \textbf{58.59}/\textbf{53.45} & \textbf{58.59}/\textbf{49.63} & \textbf{62.77}/\textbf{40.74} \\
\midrule
LSQ & 3/3 & 62.90/58.28 & 62.90/14.16 & \textbf{58.84}/\textbf{43.90} & \textbf{58.84}/14.53 & 57.48/52.36 & 57.48/47.56 &  59.74/38.47 \\
SAGM+LSQ & 3/3 & 63.00/\textbf{58.55} & 63.00/\textbf{15.01} & 57.61/43.22 & 57.61/16.39 & \textbf{59.23}/53.73 & \textbf{59.23}/49.74 & \textbf{59.95}/39.44 \\
\rowcolor{lightyellow}
Ours & 3/3 & \textbf{63.07}/58.50 & \textbf{63.07}/14.97 & 57.69/43.35 & 57.69/\textbf{16.40} & 58.68/\textbf{54.01} & 58.68/\textbf{49.97} & 59.81/\textbf{39.53} \\
\bottomrule
\end{tabularx}
\end{table}

\begin{wrapfigure}{r}{0.5\textwidth}  % 图片在右边，宽度为0.4倍的文本宽度
    \centering
    \includegraphics[width=0.45\textwidth]{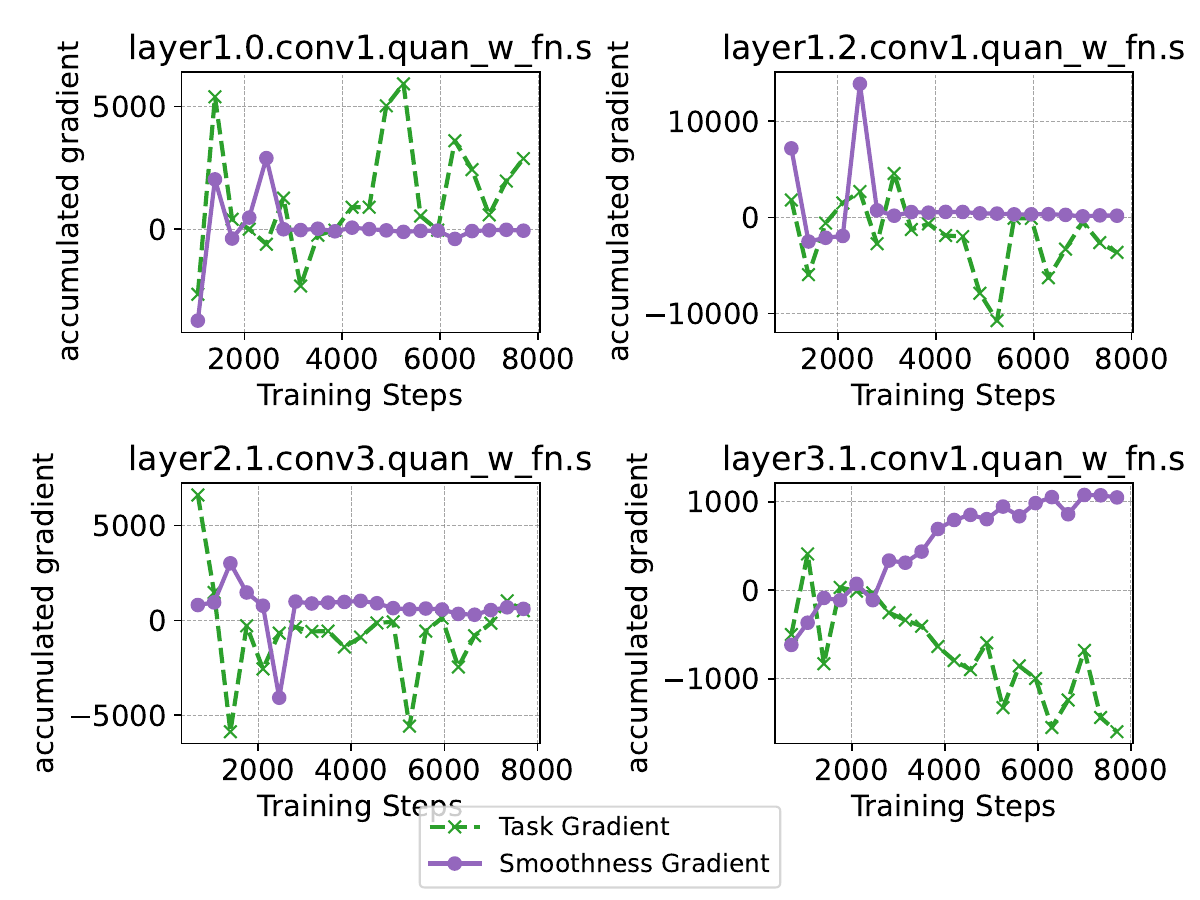} % 插入图片，调整宽度
    \caption{Results of cumulative gradients every 2111 steps in the 3-bit test on the DoaminNet Clipart and Infograph domains,
revealing fewer anomalous gradients compared to 4-bit, with \(\mathbf{g}_{{\text{task}}}\) dominating.}
    \label{fig:3bitdomain}
\end{wrapfigure}

On the DomainNet dataset, at 4-bit quantization, our method achieves an average test accuracy of \textbf{40.74\%}, surpassing both LSQ and SAGM+LSQ, and nearing the full-precision accuracy of \textbf{40.95\%}, consistently delivering the best performance across all domains. With 3-bit quantization, it achieves \textbf{39.53\%}, maintaining the best performance, though with a slight drop in validation accuracy. We observed fewer scale gradient conflicts in 3-bit compared to 4-bit (see Figure~\ref{fig:3bitdomain}), where task gradients dominate. This explains the slight validation drop when freezing task gradients, supporting the effectiveness of our approach.

\vspace{-10pt} % 调整顶部的垂直间距

\subsection{Ablation Study}

In our analysis, we validated the effectiveness of freezing \(\mathbf{g}_{{\text{task}}}\) with gradient disorder below a specific threshold and periodically reselecting the freeze set to stabilize quantization training in the DG scenario. A natural question arises: what if we reverse these choices? Specifically, what happens if we freeze scaling factors with gradient disorder above the threshold, or if we do not unfreeze after freezing?
% \begin{wrapfigure}{r}{0.7\linewidth}
%     \centering
%     \begin{subfigure}[b]{0.49\linewidth}
%         \centering
%         \includegraphics[width=\linewidth]{ICLR_2025_Template/iclr2025_pics/freeze_steps.pdf}
%         \caption{Freeze Steps Comparison}
%         \label{fig:freeze-steps}
%     \end{subfigure}
%     \hfill
%     \begin{subfigure}[b]{0.49\linewidth}
%         \centering
%         \includegraphics[width=\linewidth]{ICLR_2025_Template/iclr2025_pics/Threshold.pdf}
%         \caption{Threshold Comparison}
%         \label{fig:threshold-comparison}
%     \end{subfigure}
%     \caption{Performance comparison for Freeze Steps and Thresholds in IID and OOD domains.}
%     \label{fig:combined-figures}
% \end{wrapfigure}

As shown in Table~\ref{tab:ablation}, we fixed the freeze steps at 350 and set the threshold at 0.3 on the PACS dataset. We denote the strategy of freezing scaling factors above the threshold (with reselection) as \textit{Ours (Reverse Ratio)} and continuous freezing without unfreezing as \textit{Ours (w/o Unfreeze)}. It can be seen that simply not unfreezing still leads to a certain improvement in OOD performance. However, if we apply reverse freezing, it significantly decreases performance on both the validation and test sets. This further validating the effectiveness of our proposed method.
\begin{table}[h!]
\centering
\caption{Ablation Study on PACS: Effect of Freezing Strategies}
\label{tab:ablation}
\scriptsize
\renewcommand{\arraystretch}{1.1}
\resizebox{\textwidth}{!}{
\begin{tabular}{@{}lcccccc@{}}
\toprule
\rowcolor{lightgray}
Method & Bit-width (W/A) & Art (val/test) & Cartoon (val/test) & Photo (val/test) & Sketch (val/test) & Avg (val/test) \\
\midrule
SAGM+LSQ & 4/4 & 86.21/46.49 & 76.86/55.12 & 81.79/64.67 & 73.61/58.81 & 79.62/56.27 \\
Ours (Reverse Ratio) & 4/4 & 84.64/45.21 & 78.01/55.33 & 77.61/60.10 & 74.33/59.80 & 78.65/55.11 \\
Ours (w/o Unfreeze) & 4/4 & 86.81/\textbf{48.51} & 77.87/56.66 & 77.61/60.10 & 75.71/\textbf{63.55} & 79.5/57.21 \\
\rowcolor{lightyellow}
Ours & 4/4 & \textbf{87.45}/48.20 & \textbf{78.11}/\textbf{59.22} & \textbf{83.48}/\textbf{67.51} & \textbf{75.75}/62.37 & \textbf{81.20}/\textbf{59.33} \\
\bottomrule
\end{tabular}
}
\end{table}

Additionally, we analyzed the sensitivity of different domains to hyperparameter settings using the 4-bit configuration on PACS. We fixed the number of freeze steps and varied the threshold, as shown in Tables~\ref{tab:ablation2} and \ref{tab:ablation3}. The results indicate that different domains exhibit varying sensitivities to hyperparameters. Within a certain reasonable range, it is the level of gradient disorder threshold that ultimately determines performance, while the step size remains relatively insensitive. Therefore, establishing distinct hyperparameter search spaces for each domain could lead to improved performance.

\begin{table}[h!]
\centering
\caption{Ablation Study on PACS: Effect of Freeze Steps}
\label{tab:ablation2}
\scriptsize
\renewcommand{\arraystretch}{1.1}
\resizebox{\textwidth}{!}{
\begin{tabular}{@{}lcccccc@{}}
\toprule
\rowcolor{lightgray}
Freeze Steps & Bit-width (W/A) & Art (val/test) & Cartoon (val/test) & Photo (val/test) & Sketch (val/test) & Avg (val/test) \\
\midrule
300 & 4/4 & 85.85/47.65 & 78.05/58.37 & \textbf{84.32}/69.09 & 74.35/61.80 & 80.64/59.23 \\
350 & 4/4 & \textbf{87.45}/48.20 & 78.11/\textbf{59.22} & 83.48/67.51 & 75.75/62.37 & \textbf{81.20}/\textbf{59.33} \\
400 & 4/4 & 86.51/\textbf{48.38} & \textbf{78.44}/56.66 & 82.77/\textbf{69.09} & \textbf{76.67}/\textbf{62.53} & 81.10/59.17 \\
\bottomrule
\end{tabular}
}
\end{table}

\begin{table}[h!]
\centering
\caption{Ablation Study on PACS: Effect of Threshold \(r\)}
\label{tab:ablation3}
\scriptsize
\renewcommand{\arraystretch}{1.1}
\resizebox{\textwidth}{!}{
\begin{tabular}{@{}lcccccc@{}}
\toprule
\rowcolor{lightgray}
Threshold \(r\) & Bit-width (W/A) & Art (val/test) & Cartoon (val/test) & Photo (val/test) & Sketch (val/test) & Avg (val/test) \\
\midrule
0.28 & 4/4 & 86.74/\textbf{49.24} & 77.79/55.92 & 79.77/64.22 & 74.59/62.21 & 79.72/57.90 \\
0.30 & 4/4 & \textbf{87.45}/48.20 & \textbf{77.87}/\textbf{56.45} & 79.46/63.62 & 75.75/62.37 & 80.13/57.66 \\
0.32 & 4/4 & 86.96/48.63 & 77.20/55.17 & \textbf{80.62}/\textbf{64.60} & \textbf{77.25}/\textbf{67.40} & \textbf{80.51}/\textbf{58.95} \\
\bottomrule
\end{tabular}
}
\end{table}

\subsection{\textcolor{black}{Loss surface visualization}}
\begin{figure}[htbp]
    \centering
    % First row: bad
    \begin{subfigure}[b]{0.23\linewidth}
        \centering
        \includegraphics[width=\linewidth]{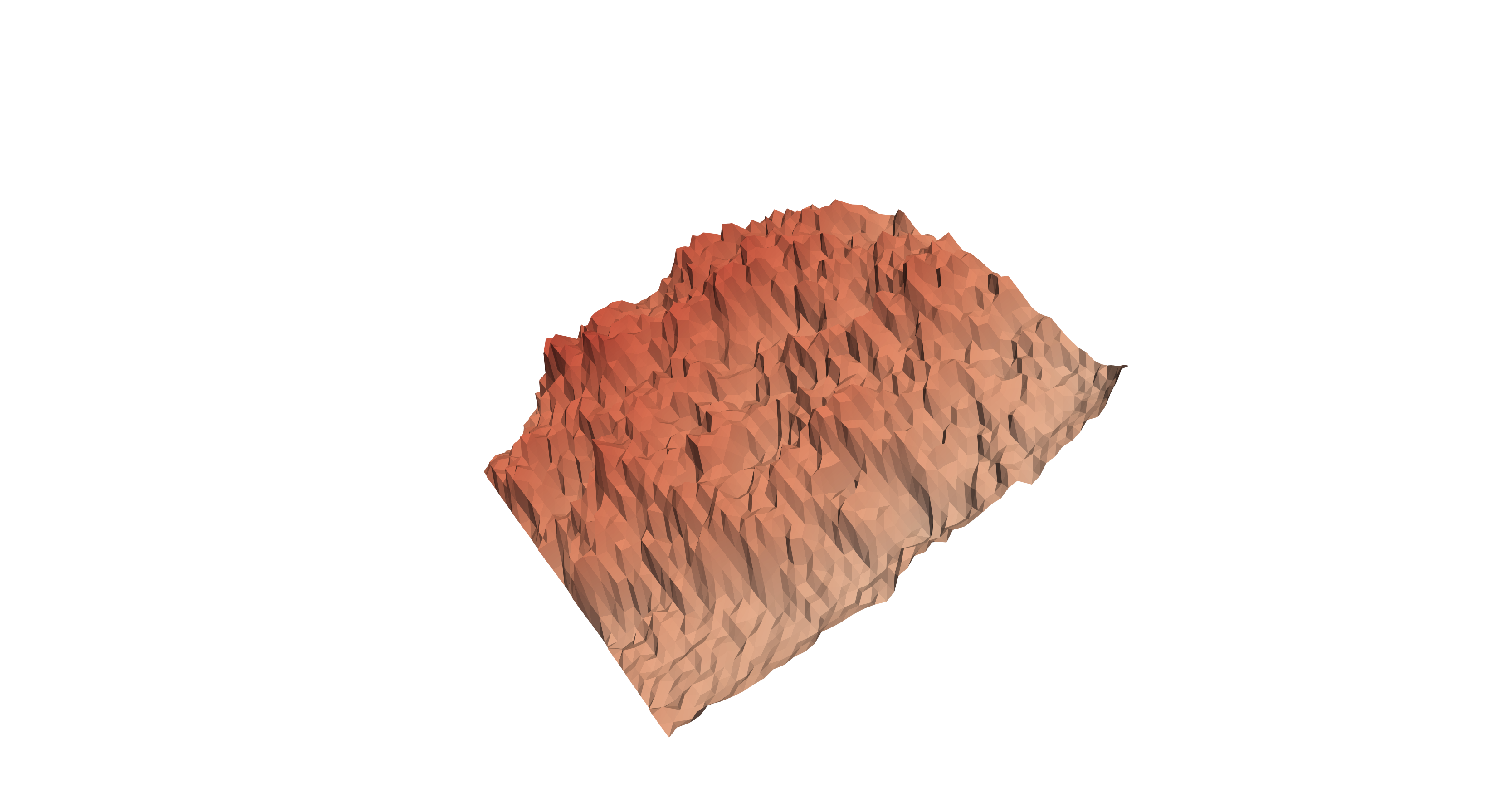}
        \caption{\textcolor{black}{Art(Origin)}}
        \label{fig:bad0}
    \end{subfigure}
    \hfill
    \begin{subfigure}[b]{0.23\linewidth}
        \centering
        \includegraphics[width=\linewidth]{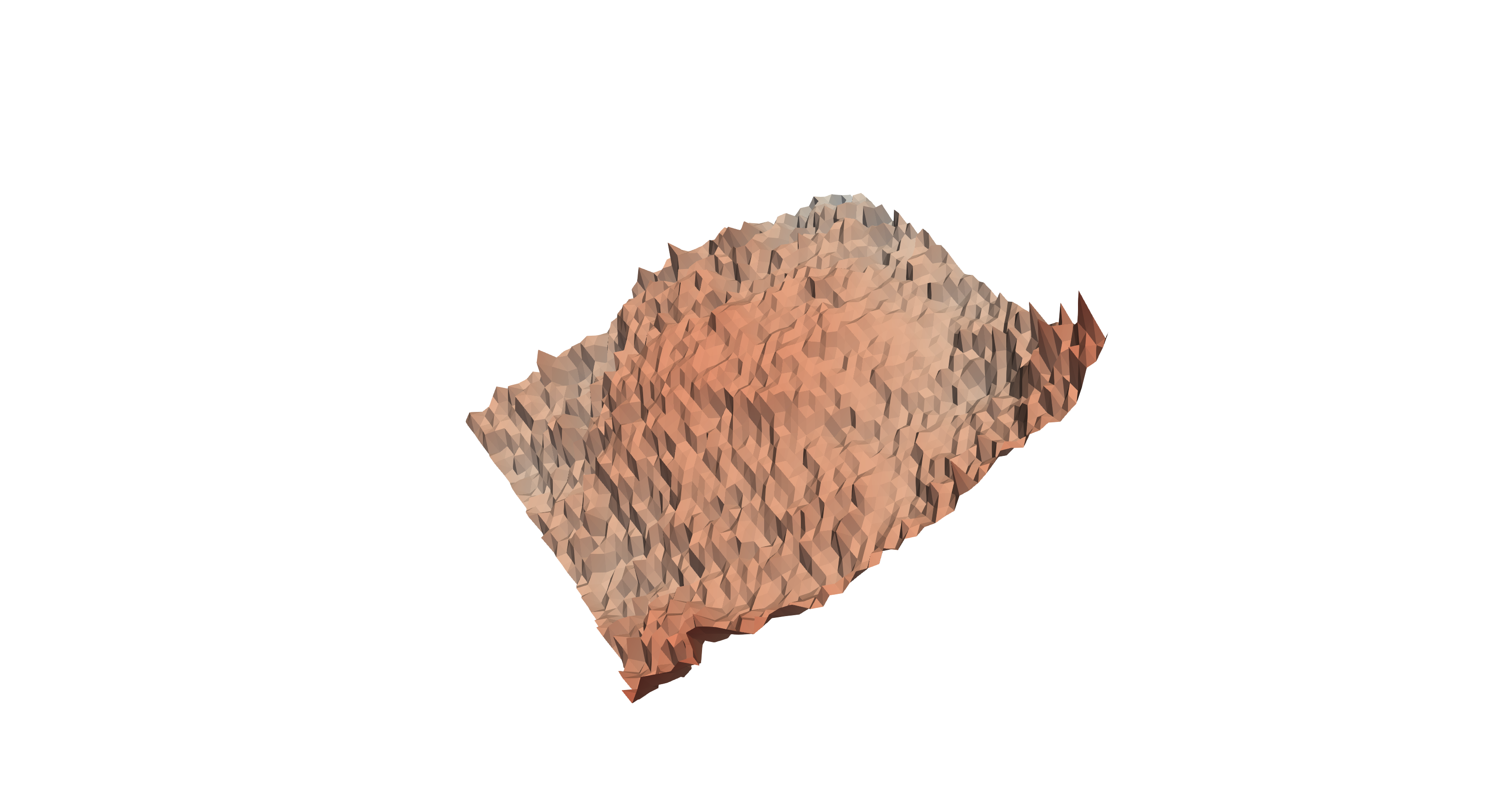}
        \caption{\textcolor{black}{Cartoon(Origin)}}
        \label{fig:bad1}
    \end{subfigure}
    \hfill
    \begin{subfigure}[b]{0.23\linewidth}
        \centering
        \includegraphics[width=\linewidth]{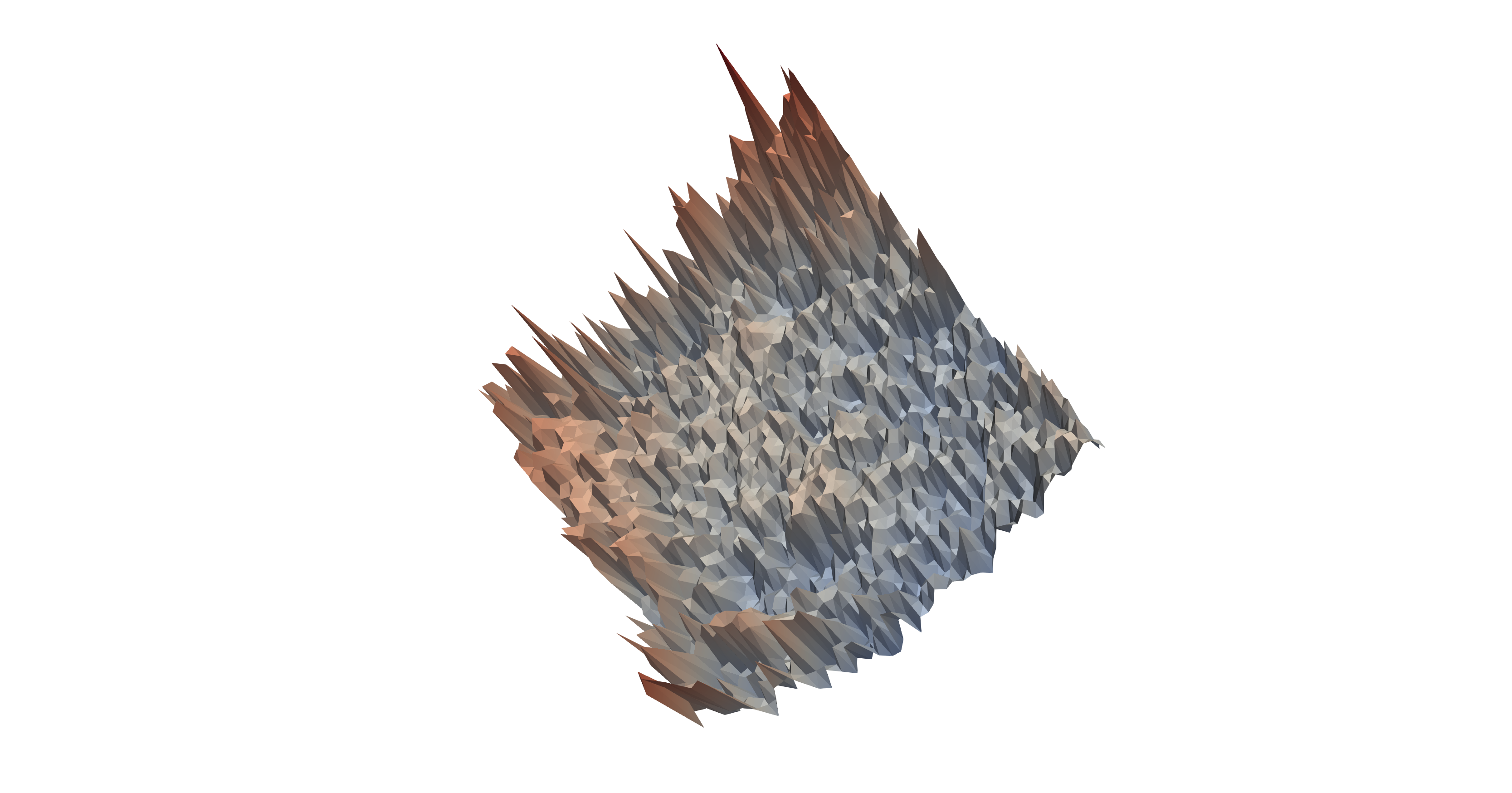}
        \caption{\textcolor{black}{Photo(Origin)}}
        \label{fig:bad3}
    \end{subfigure}
    \hfill
    \begin{subfigure}[b]{0.23\linewidth}
        \centering
        \includegraphics[width=\linewidth]{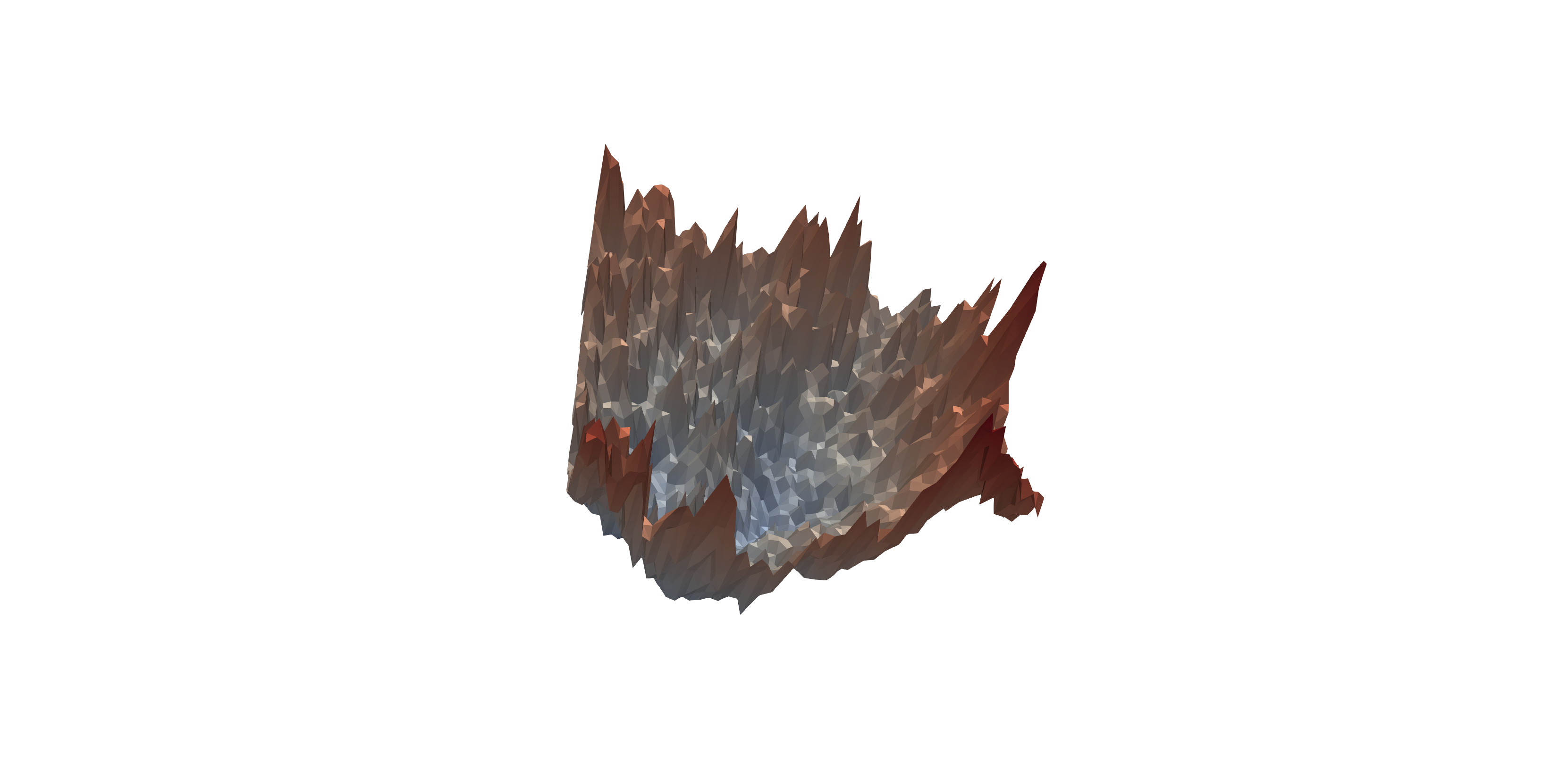}
        \caption{\textcolor{black}{Sketch(Origin)}}
        \label{fig:origin}
    \end{subfigure}
    
    \vspace{1em} % Add vertical space between rows

    % Second row: good
    \begin{subfigure}[b]{0.23\linewidth}
        \centering
        \includegraphics[width=\linewidth]{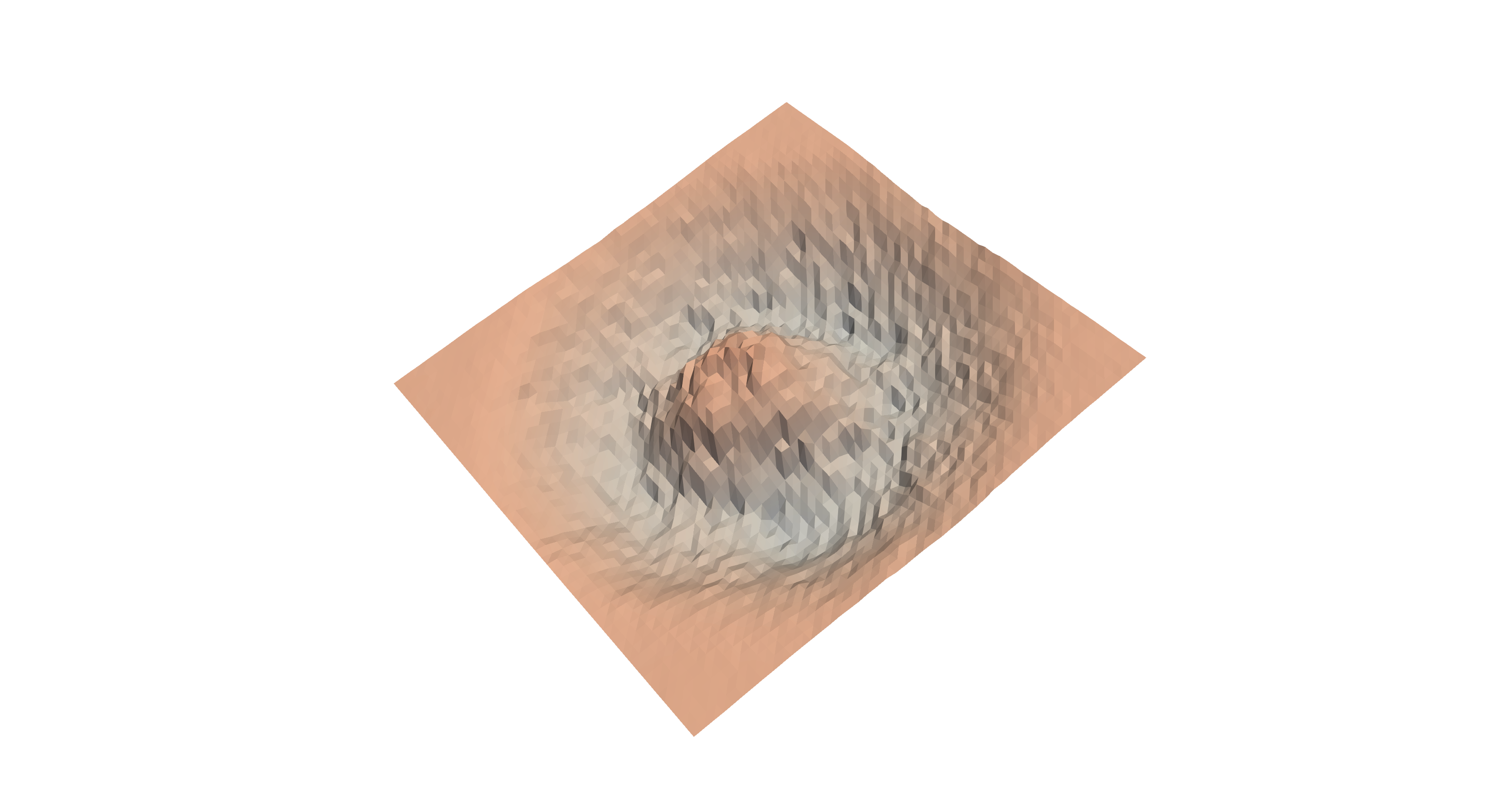}
        \caption{\textcolor{black}{Art(Ours)}}
        \label{fig:good0}
    \end{subfigure}
    \hfill
    \begin{subfigure}[b]{0.23\linewidth}
        \centering
        \includegraphics[width=\linewidth]{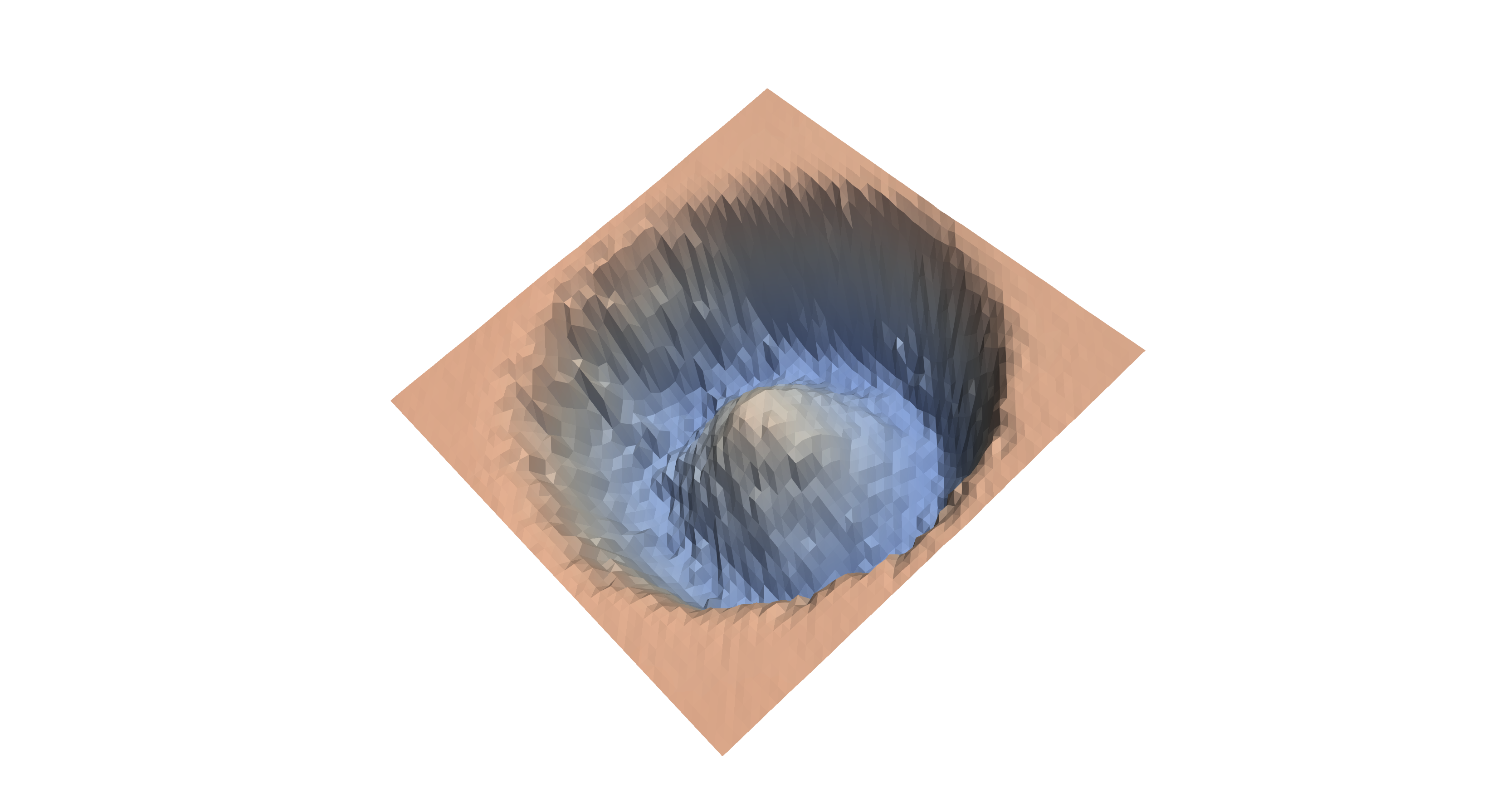}
        \caption{\textcolor{black}{Cartoon(Ours)}}
        \label{fig:good1}
    \end{subfigure}
    \hfill
    \begin{subfigure}[b]{0.23\linewidth}
        \centering
        \includegraphics[width=\linewidth]{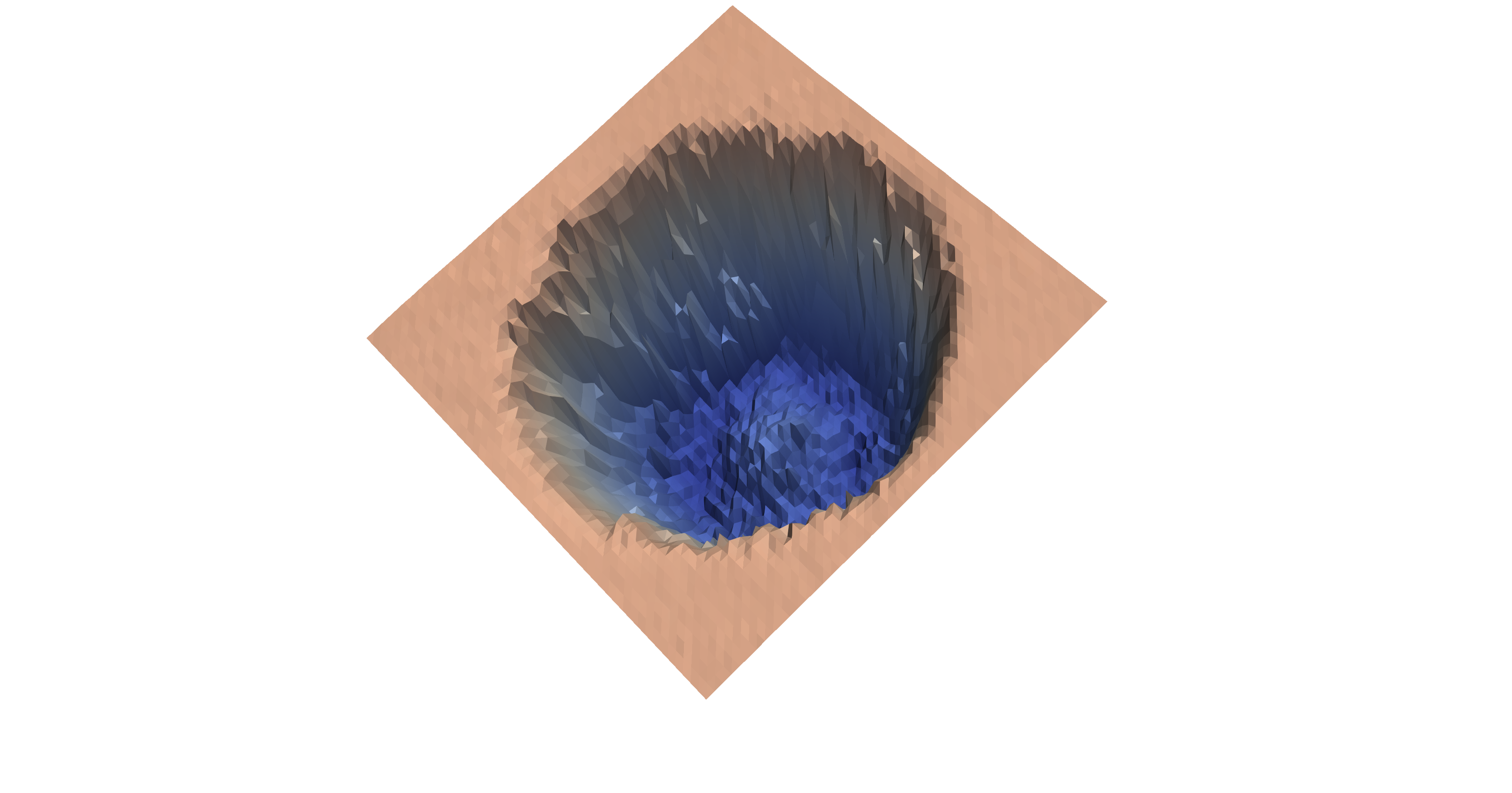}
        \caption{\textcolor{black}{Photo(Ours)}}
        \label{fig:good3}
    \end{subfigure}
    \hfill
    \begin{subfigure}[b]{0.23\linewidth}
        \centering
        \includegraphics[width=\linewidth]{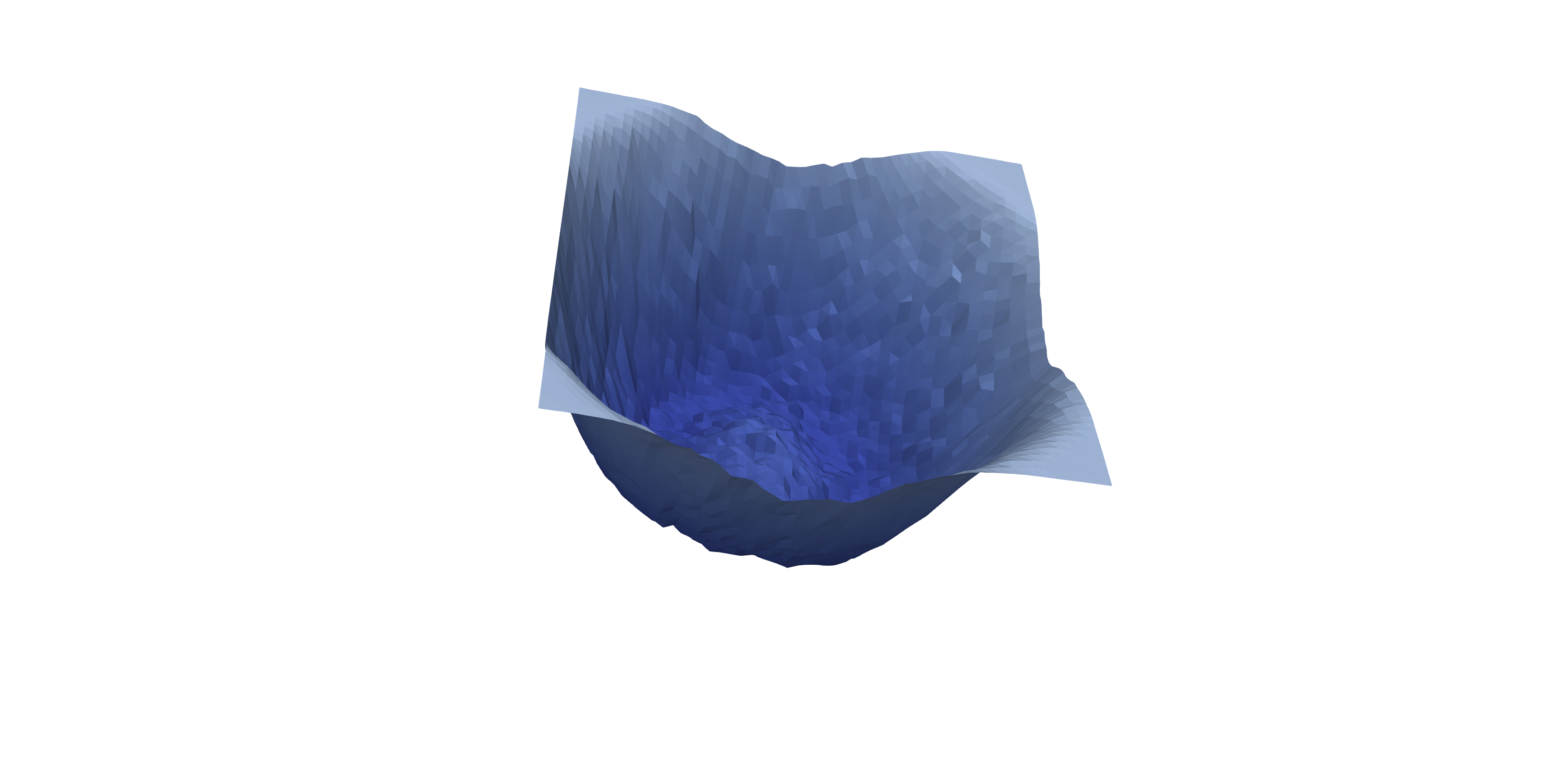}
        \caption{\textcolor{black}{Sketch(Ours)}}
        \label{fig:smooth}
    \end{subfigure}

    \caption{\textcolor{black}{Visualization of the loss landscape across various domains. Top is the direct integration of SAGM into QAT, bottom is proposed method. Our method achieves smoother loss surfaces across all four domains in PACS.}}
    \label{fig:bad_good_comparison}
\end{figure}

% 图x显示了在PACS四个域上，直接引入sagm与采用我们的方法之后的loss平面可视化差异，可以直观看出在PACS四个域上，我们方法获得的模型均显著更平滑。
\textcolor{black}{Following the approach in \citep{li2018visualizing}, Figure \ref{fig:bad_good_comparison} illustrates the differences in loss surface visualizations across the four domains of PACS when incorporating SAGM directly versus applying our proposed method. The results clearly show that our method consistently achieves significantly smoother loss surfaces across all four domains.}

%% file: related_work.tex
\section{Related Work}

\subsection{Domain Generalization}
In practical applications, when deploying machine learning models, test data distribution may differ from the training distribution, a common phenomenon known as distribution shift~\citep{liu2021towards,yu2024survey,koh2021wilds}. Domain generalization (DG) aims to enhance a model's ability to generalize to unseen domains~\citep{wang2022generalizing, zhou2022domain}. Common strategies include domain alignment ~\citep{muandet2013domain, li2018domain,zhao2020domain}, meta learning~\citep{li2018learning,balaji2018metareg,dou2019domain}, data augmentation ~\citep{zhou2021domain,carlucci2019domain}, disentangled representation learning ~\citep{zhang2022towards} and utilization of causal relations~\citep{mahajan2021domain,lv2022causality}. 
Inspired by previous studies of flat minima~\citep{izmailov2018averaging,foret2020sharpness,liu2022towards,zhuang2022surrogate,zhang2023gradient}, flatness-aware methods start to gain attention and exhibit remarkable performance in domain generalization \citep{cha2021swad,wang2023sharpness,zhang2023flatness}, such as SAGM\citep{wang2023sharpness}, which improves generalization by optimizing the angle between weight gradients. However, these methods primarily focus on full-precision models, which are impractical for deployment on edge devices commonly used in high-risk scenarios and do not take into account the factors specific to quantization. We specifically focus on strategies to enhance model generalization in quantized training environments.

\subsection{Quantizaion-aware Training} 
Quantizaion-aware training (QAT) involves inserting simulated quantization nodes and retraining the model, which achieves a better balance between accuracy and compression ratio~\citep{hubara2021accurate,pmlr-v119-nagel20a}. 
% This paper will focus on QAT methods. Due to the truncation characteristics of quantization, QAT typically employs the Straight-Through Estimator (STE)~\citep{bengio2013estimating} to approximate gradients, which is the main source of quantization error. 
DoReFa~\citep{zhou2016dorefa} and PACT~\citep{choi2018pact} use low-precision weights and activations during the forward pass and utilize STE techniques~\citep{bengio2013estimating} during backpropagation to estimate gradients of the piece-wise quantization functions. LSQ~\citep{esser2019learned} adjusts the quantization function by introducing learnable step size scaling factors. 
Recently, some works have explored the possibility of improving quantization performance by freezing unstable weights to further enhance results~\citep{nagel2022overcoming,tang2024retraining,liu2023oscillation}; however, these methods have only considered the Identically Distributed (I.I.D.) assumptions. 
Due to distribution shifts in unseen data—which often occur in practical applications—the quality and reliability of quantized models cannot be guaranteed~\citep{hu2022characterizing}. 
% Compared with these studies, we focus more on the generalization performance of QAT in Out-of-Distribution (OOD) scenarios, aiming to explore QAT methods suitable for OOD environments.

%% file: CONCLUSION_AND_FUTURE_WORK.tex
\section{CONCLUSION AND FUTURE WORK}

In this paper, we propose GAQAT for domain generalization. We introduce a smoothing factor into the quantizer to jointly optimize quantization and smoothness. Our analysis of quantizer gradients revealed significant conflicts between task loss and smoothness loss due to gradient approximations, impacting generalization. To address this, we define gradient disorder to quantify quantizer gradient conflicts and designed a dynamic freezing strategy that selectively updates quantizers based on disorder levels, ensuring global performance convergence. Extensive experiments on PACS and DomainNet, along with ablation studies, demonstrate the effectiveness of GAQAT.

\textbf{Limitations and future work.}
Although we incorporated SAGM's smoothing objective into quantization, other smoothing objectives may also impact scaling factor gradients, suggesting future research potential. Our experiments reveal varying domain sensitivity to scaling factor gradients, but we only examined conflicts between task and flatness objectives. The relationship between domains and scaling factors remains unexplored.

%% file: iclr2025_conference.bbl
\begin{thebibliography}{53}
\providecommand{\natexlab}[1]{#1}
\providecommand{\url}[1]{\texttt{#1}}
\expandafter\ifx\csname urlstyle\endcsname\relax
  \providecommand{\doi}[1]{doi: #1}\else
  \providecommand{\doi}{doi: \begingroup \urlstyle{rm}\Url}\fi

\bibitem[Andriushchenko \& Flammarion(2022)Andriushchenko and Flammarion]{andriushchenko2022towards}
Maksym Andriushchenko and Nicolas Flammarion.
\newblock Towards understanding sharpness-aware minimization.
\newblock In \emph{International Conference on Machine Learning}, pp.\  639--668. PMLR, 2022.

\bibitem[Balaji et~al.(2018)Balaji, Sankaranarayanan, and Chellappa]{balaji2018metareg}
Yogesh Balaji, Swami Sankaranarayanan, and Rama Chellappa.
\newblock Metareg: Towards domain generalization using meta-regularization.
\newblock \emph{Advances in neural information processing systems}, 31, 2018.

\bibitem[Bengio et~al.(2013)Bengio, L{\'e}onard, and Courville]{bengio2013estimating}
Yoshua Bengio, Nicholas L{\'e}onard, and Aaron Courville.
\newblock Estimating or propagating gradients through stochastic neurons for conditional computation.
\newblock \emph{arXiv preprint arXiv:1308.3432}, 2013.

\bibitem[Carlucci et~al.(2019)Carlucci, D'Innocente, Bucci, Caputo, and Tommasi]{carlucci2019domain}
Fabio~M Carlucci, Antonio D'Innocente, Silvia Bucci, Barbara Caputo, and Tatiana Tommasi.
\newblock Domain generalization by solving jigsaw puzzles.
\newblock In \emph{Proceedings of the IEEE/CVF conference on computer vision and pattern recognition}, pp.\  2229--2238, 2019.

\bibitem[Cha et~al.(2021)Cha, Chun, Lee, Cho, Park, Lee, and Park]{cha2021swad}
Junbum Cha, Sanghyuk Chun, Kyungjae Lee, Han-Cheol Cho, Seunghyun Park, Yunsung Lee, and Sungrae Park.
\newblock Swad: Domain generalization by seeking flat minima.
\newblock \emph{Advances in Neural Information Processing Systems}, 34:\penalty0 22405--22418, 2021.

\bibitem[Chen et~al.(2020)Chen, Fan, Girshick, and He]{chen2020improvedbaselinesmomentumcontrastive}
Xinlei Chen, Haoqi Fan, Ross Girshick, and Kaiming He.
\newblock Improved baselines with momentum contrastive learning, 2020.
\newblock URL \url{https://arxiv.org/abs/2003.04297}.

\bibitem[Choi et~al.(2018)Choi, Wang, Venkataramani, Chuang, Srinivasan, and Gopalakrishnan]{choi2018pact}
Jungwook Choi, Zhuo Wang, Swagath Venkataramani, Pierce I-Jen Chuang, Vijayalakshmi Srinivasan, and Kailash Gopalakrishnan.
\newblock Pact: Parameterized clipping activation for quantized neural networks.
\newblock \emph{arXiv preprint arXiv:1805.06085}, 2018.

\bibitem[Dosovitskiy(2020)]{dosovitskiy2020image}
Alexey Dosovitskiy.
\newblock An image is worth 16x16 words: Transformers for image recognition at scale.
\newblock \emph{arXiv preprint arXiv:2010.11929}, 2020.

\bibitem[Dou et~al.(2019)Dou, Coelho~de Castro, Kamnitsas, and Glocker]{dou2019domain}
Qi~Dou, Daniel Coelho~de Castro, Konstantinos Kamnitsas, and Ben Glocker.
\newblock Domain generalization via model-agnostic learning of semantic features.
\newblock \emph{Advances in neural information processing systems}, 32, 2019.

\bibitem[Esser et~al.(2019)Esser, McKinstry, Bablani, Appuswamy, and Modha]{esser2019learned}
Steven~K Esser, Jeffrey~L McKinstry, Deepika Bablani, Rathinakumar Appuswamy, and Dharmendra~S Modha.
\newblock Learned step size quantization.
\newblock \emph{arXiv preprint arXiv:1902.08153}, 2019.

\bibitem[Foret et~al.(2020)Foret, Kleiner, Mobahi, and Neyshabur]{foret2020sharpness}
Pierre Foret, Ariel Kleiner, Hossein Mobahi, and Behnam Neyshabur.
\newblock Sharpness-aware minimization for efficiently improving generalization.
\newblock \emph{arXiv preprint arXiv:2010.01412}, 2020.

\bibitem[Gulrajani \& Lopez-Paz(2020)Gulrajani and Lopez-Paz]{gulrajani2020search}
Ishaan Gulrajani and David Lopez-Paz.
\newblock In search of lost domain generalization.
\newblock \emph{arXiv preprint arXiv:2007.01434}, 2020.

\bibitem[He et~al.(2016)He, Zhang, Ren, and Sun]{he2016deep}
Kaiming He, Xiangyu Zhang, Shaoqing Ren, and Jian Sun.
\newblock Deep residual learning for image recognition.
\newblock In \emph{Proceedings of the IEEE conference on computer vision and pattern recognition}, pp.\  770--778, 2016.

\bibitem[Hu et~al.(2022)Hu, Guo, Cordy, Xie, Ma, Papadakis, and Traon]{hu2022characterizing}
Qiang Hu, Yuejun Guo, Maxime Cordy, Xiaofei Xie, Wei Ma, Mike Papadakis, and Yves~Le Traon.
\newblock Characterizing and understanding the behavior of quantized models for reliable deployment.
\newblock \emph{arXiv preprint arXiv:2204.04220}, 2022.

\bibitem[Hubara et~al.(2021)Hubara, Nahshan, Hanani, Banner, and Soudry]{hubara2021accurate}
Itay Hubara, Yury Nahshan, Yair Hanani, Ron Banner, and Daniel Soudry.
\newblock Accurate post training quantization with small calibration sets.
\newblock In \emph{International Conference on Machine Learning}, pp.\  4466--4475. PMLR, 2021.

\bibitem[Izmailov et~al.(2018)Izmailov, Podoprikhin, Garipov, Vetrov, and Wilson]{izmailov2018averaging}
Pavel Izmailov, Dmitrii Podoprikhin, Timur Garipov, Dmitry Vetrov, and Andrew~Gordon Wilson.
\newblock Averaging weights leads to wider optima and better generalization.
\newblock \emph{arXiv preprint arXiv:1803.05407}, 2018.

\bibitem[Koh et~al.(2021)Koh, Sagawa, Marklund, Xie, Zhang, Balsubramani, Hu, Yasunaga, Phillips, Gao, et~al.]{koh2021wilds}
Pang~Wei Koh, Shiori Sagawa, Henrik Marklund, Sang~Michael Xie, Marvin Zhang, Akshay Balsubramani, Weihua Hu, Michihiro Yasunaga, Richard~Lanas Phillips, Irena Gao, et~al.
\newblock Wilds: A benchmark of in-the-wild distribution shifts.
\newblock In \emph{International conference on machine learning}, pp.\  5637--5664. PMLR, 2021.

\bibitem[Li \& Giannakis(2024)Li and Giannakis]{li2024enhancing}
Bingcong Li and Georgios Giannakis.
\newblock Enhancing sharpness-aware optimization through variance suppression.
\newblock \emph{Advances in Neural Information Processing Systems}, 36, 2024.

\bibitem[Li et~al.(2017)Li, Yang, Song, and Hospedales]{Li_2017_ICCV}
Da~Li, Yongxin Yang, Yi-Zhe Song, and Timothy~M. Hospedales.
\newblock Deeper, broader and artier domain generalization.
\newblock In \emph{Proceedings of the IEEE International Conference on Computer Vision (ICCV)}, Oct 2017.

\bibitem[Li et~al.(2018{\natexlab{a}})Li, Yang, Song, and Hospedales]{li2018learning}
Da~Li, Yongxin Yang, Yi-Zhe Song, and Timothy Hospedales.
\newblock Learning to generalize: Meta-learning for domain generalization.
\newblock In \emph{Proceedings of the AAAI conference on artificial intelligence}, volume~32, 2018{\natexlab{a}}.

\bibitem[Li et~al.(2018{\natexlab{b}})Li, Xu, Taylor, Studer, and Goldstein]{li2018visualizing}
Hao Li, Zheng Xu, Gavin Taylor, Christoph Studer, and Tom Goldstein.
\newblock Visualizing the loss landscape of neural nets.
\newblock \emph{Advances in neural information processing systems}, 31, 2018{\natexlab{b}}.

\bibitem[Li et~al.(2018{\natexlab{c}})Li, Gong, Tian, Liu, and Tao]{li2018domain}
Ya~Li, Mingming Gong, Xinmei Tian, Tongliang Liu, and Dacheng Tao.
\newblock Domain generalization via conditional invariant representations.
\newblock In \emph{Proceedings of the AAAI conference on artificial intelligence}, volume~32, 2018{\natexlab{c}}.

\bibitem[Liu et~al.(2021)Liu, Shen, He, Zhang, Xu, Yu, and Cui]{liu2021towards}
Jiashuo Liu, Zheyan Shen, Yue He, Xingxuan Zhang, Renzhe Xu, Han Yu, and Peng Cui.
\newblock Towards out-of-distribution generalization: A survey.
\newblock \emph{arXiv preprint arXiv:2108.13624}, 2021.

\bibitem[Liu et~al.(2023)Liu, Liu, and Cheng]{liu2023oscillation}
Shih-Yang Liu, Zechun Liu, and Kwang-Ting Cheng.
\newblock Oscillation-free quantization for low-bit vision transformers.
\newblock In \emph{International Conference on Machine Learning}, pp.\  21813--21824. PMLR, 2023.

\bibitem[Liu et~al.(2022)Liu, Mai, Chen, Hsieh, and You]{liu2022towards}
Yong Liu, Siqi Mai, Xiangning Chen, Cho-Jui Hsieh, and Yang You.
\newblock Towards efficient and scalable sharpness-aware minimization.
\newblock In \emph{Proceedings of the IEEE/CVF Conference on Computer Vision and Pattern Recognition}, pp.\  12360--12370, 2022.

\bibitem[Lv et~al.(2022)Lv, Liang, Li, Zang, Liu, Wang, and Liu]{lv2022causality}
Fangrui Lv, Jian Liang, Shuang Li, Bin Zang, Chi~Harold Liu, Ziteng Wang, and Di~Liu.
\newblock Causality inspired representation learning for domain generalization.
\newblock In \emph{Proceedings of the IEEE/CVF conference on computer vision and pattern recognition}, pp.\  8046--8056, 2022.

\bibitem[Mahajan et~al.(2021)Mahajan, Tople, and Sharma]{mahajan2021domain}
Divyat Mahajan, Shruti Tople, and Amit Sharma.
\newblock Domain generalization using causal matching.
\newblock In \emph{International conference on machine learning}, pp.\  7313--7324. PMLR, 2021.

\bibitem[Muandet et~al.(2013)Muandet, Balduzzi, and Sch{\"o}lkopf]{muandet2013domain}
Krikamol Muandet, David Balduzzi, and Bernhard Sch{\"o}lkopf.
\newblock Domain generalization via invariant feature representation.
\newblock In \emph{International conference on machine learning}, pp.\  10--18. PMLR, 2013.

\bibitem[Nagel et~al.(2020)Nagel, Amjad, Van~Baalen, Louizos, and Blankevoort]{pmlr-v119-nagel20a}
Markus Nagel, Rana~Ali Amjad, Mart Van~Baalen, Christos Louizos, and Tijmen Blankevoort.
\newblock Up or down? {A}daptive rounding for post-training quantization.
\newblock In Hal~Daumé III and Aarti Singh (eds.), \emph{Proceedings of the 37th International Conference on Machine Learning}, volume 119 of \emph{Proceedings of Machine Learning Research}, pp.\  7197--7206. PMLR, 13--18 Jul 2020.
\newblock URL \url{https://proceedings.mlr.press/v119/nagel20a.html}.

\bibitem[Nagel et~al.(2021)Nagel, Fournarakis, Amjad, Bondarenko, van Baalen, and Blankevoort]{nagel2021whitepaperneuralnetwork}
Markus Nagel, Marios Fournarakis, Rana~Ali Amjad, Yelysei Bondarenko, Mart van Baalen, and Tijmen Blankevoort.
\newblock A white paper on neural network quantization, 2021.
\newblock URL \url{https://arxiv.org/abs/2106.08295}.

\bibitem[Nagel et~al.(2022)Nagel, Fournarakis, Bondarenko, and Blankevoort]{nagel2022overcoming}
Markus Nagel, Marios Fournarakis, Yelysei Bondarenko, and Tijmen Blankevoort.
\newblock Overcoming oscillations in quantization-aware training.
\newblock In \emph{International Conference on Machine Learning}, pp.\  16318--16330. PMLR, 2022.

\bibitem[Peng et~al.(2019)Peng, Huang, Sun, and Saenko]{peng2019domain}
Xingchao Peng, Zijun Huang, Ximeng Sun, and Kate Saenko.
\newblock Domain agnostic learning with disentangled representations.
\newblock In \emph{International conference on machine learning}, pp.\  5102--5112. PMLR, 2019.

\bibitem[Sandler et~al.(2018)Sandler, Howard, Zhu, Zhmoginov, and Chen]{sandler2018mobilenetv2}
Mark Sandler, Andrew Howard, Menglong Zhu, Andrey Zhmoginov, and Liang-Chieh Chen.
\newblock Mobilenetv2: Inverted residuals and linear bottlenecks.
\newblock In \emph{Proceedings of the IEEE conference on computer vision and pattern recognition}, pp.\  4510--4520, 2018.

\bibitem[Strudel et~al.(2021)Strudel, Garcia, Laptev, and Schmid]{strudel2021segmenter}
Robin Strudel, Ricardo Garcia, Ivan Laptev, and Cordelia Schmid.
\newblock Segmenter: Transformer for semantic segmentation.
\newblock In \emph{Proceedings of the IEEE/CVF international conference on computer vision}, pp.\  7262--7272, 2021.

\bibitem[Tang et~al.(2022)Tang, Ouyang, Wang, Zhu, Ji, Wang, and Zhu]{tang2022mixed}
Chen Tang, Kai Ouyang, Zhi Wang, Yifei Zhu, Wen Ji, Yaowei Wang, and Wenwu Zhu.
\newblock Mixed-precision neural network quantization via learned layer-wise importance.
\newblock In \emph{European Conference on Computer Vision}, pp.\  259--275. Springer, 2022.

\bibitem[Tang et~al.(2024)Tang, Meng, Jiang, Xie, Lu, Ma, Wang, and Zhu]{tang2024retraining}
Chen Tang, Yuan Meng, Jiacheng Jiang, Shuzhao Xie, Rongwei Lu, Xinzhu Ma, Zhi Wang, and Wenwu Zhu.
\newblock Retraining-free model quantization via one-shot weight-coupling learning.
\newblock In \emph{Proceedings of the IEEE/CVF Conference on Computer Vision and Pattern Recognition}, pp.\  15855--15865, 2024.

\bibitem[Volpi et~al.(2018)Volpi, Namkoong, Sener, Duchi, Murino, and Savarese]{volpi2018generalizing}
Riccardo Volpi, Hongseok Namkoong, Ozan Sener, John~C Duchi, Vittorio Murino, and Silvio Savarese.
\newblock Generalizing to unseen domains via adversarial data augmentation.
\newblock \emph{Advances in neural information processing systems}, 31, 2018.

\bibitem[Wang et~al.(2022)Wang, Lan, Liu, Ouyang, Qin, Lu, Chen, Zeng, and Philip]{wang2022generalizing}
Jindong Wang, Cuiling Lan, Chang Liu, Yidong Ouyang, Tao Qin, Wang Lu, Yiqiang Chen, Wenjun Zeng, and S~Yu Philip.
\newblock Generalizing to unseen domains: A survey on domain generalization.
\newblock \emph{IEEE transactions on knowledge and data engineering}, 35\penalty0 (8):\penalty0 8052--8072, 2022.

\bibitem[Wang et~al.(2023)Wang, Zhang, Lei, and Zhang]{wang2023sharpness}
Pengfei Wang, Zhaoxiang Zhang, Zhen Lei, and Lei Zhang.
\newblock Sharpness-aware gradient matching for domain generalization.
\newblock In \emph{Proceedings of the IEEE/CVF Conference on Computer Vision and Pattern Recognition}, pp.\  3769--3778, 2023.

\bibitem[Wen et~al.(2023)Wen, Ma, and Li]{wen2023sharpness}
Kaiyue Wen, Tengyu Ma, and Zhiyuan Li.
\newblock How sharpness-aware minimization minimizes sharpness?
\newblock In \emph{The Eleventh International Conference on Learning Representations}, 2023.

\bibitem[Yu et~al.(2024{\natexlab{a}})Yu, Liu, Zhang, Wu, and Cui]{yu2024survey}
Han Yu, Jiashuo Liu, Xingxuan Zhang, Jiayun Wu, and Peng Cui.
\newblock A survey on evaluation of out-of-distribution generalization.
\newblock \emph{arXiv preprint arXiv:2403.01874}, 2024{\natexlab{a}}.

\bibitem[Yu et~al.(2024{\natexlab{b}})Yu, Zhang, Xu, Liu, He, and Cui]{yu2024rethinking}
Han Yu, Xingxuan Zhang, Renzhe Xu, Jiashuo Liu, Yue He, and Peng Cui.
\newblock Rethinking the evaluation protocol of domain generalization.
\newblock In \emph{Proceedings of the IEEE/CVF Conference on Computer Vision and Pattern Recognition}, pp.\  21897--21908, 2024{\natexlab{b}}.

\bibitem[Zhang et~al.(2022{\natexlab{a}})Zhang, Zhang, Liu, Weller, Sch{\"o}lkopf, and Xing]{zhang2022towards}
Hanlin Zhang, Yi-Fan Zhang, Weiyang Liu, Adrian Weller, Bernhard Sch{\"o}lkopf, and Eric~P Xing.
\newblock Towards principled disentanglement for domain generalization.
\newblock In \emph{Proceedings of the IEEE/CVF conference on computer vision and pattern recognition}, pp.\  8024--8034, 2022{\natexlab{a}}.

\bibitem[Zhang et~al.(2022{\natexlab{b}})Zhang, Li, Liu, Zhang, Su, Zhu, Ni, and Shum]{zhang2022dino}
Hao Zhang, Feng Li, Shilong Liu, Lei Zhang, Hang Su, Jun Zhu, Lionel~M Ni, and Heung-Yeung Shum.
\newblock Dino: Detr with improved denoising anchor boxes for end-to-end object detection.
\newblock \emph{arXiv preprint arXiv:2203.03605}, 2022{\natexlab{b}}.

\bibitem[Zhang et~al.(2023{\natexlab{a}})Zhang, Xu, Yu, Dong, Tian, and Cui]{zhang2023flatness}
Xingxuan Zhang, Renzhe Xu, Han Yu, Yancheng Dong, Pengfei Tian, and Peng Cui.
\newblock Flatness-aware minimization for domain generalization.
\newblock In \emph{Proceedings of the IEEE/CVF International Conference on Computer Vision}, pp.\  5189--5202, 2023{\natexlab{a}}.

\bibitem[Zhang et~al.(2023{\natexlab{b}})Zhang, Xu, Yu, Zou, and Cui]{zhang2023gradient}
Xingxuan Zhang, Renzhe Xu, Han Yu, Hao Zou, and Peng Cui.
\newblock Gradient norm aware minimization seeks first-order flatness and improves generalization.
\newblock In \emph{Proceedings of the IEEE/CVF Conference on Computer Vision and Pattern Recognition}, pp.\  20247--20257, 2023{\natexlab{b}}.

\bibitem[Zhao et~al.(2020)Zhao, Gong, Liu, Fu, and Tao]{zhao2020domain}
Shanshan Zhao, Mingming Gong, Tongliang Liu, Huan Fu, and Dacheng Tao.
\newblock Domain generalization via entropy regularization.
\newblock \emph{Advances in neural information processing systems}, 33:\penalty0 16096--16107, 2020.

\bibitem[Zhou et~al.(2021)Zhou, Yang, Qiao, and Xiang]{zhou2021domain}
Kaiyang Zhou, Yongxin Yang, Yu~Qiao, and Tao Xiang.
\newblock Domain generalization with mixstyle.
\newblock \emph{arXiv preprint arXiv:2104.02008}, 2021.

\bibitem[Zhou et~al.(2022{\natexlab{a}})Zhou, Liu, Qiao, Xiang, and Loy]{zhou2022domain}
Kaiyang Zhou, Ziwei Liu, Yu~Qiao, Tao Xiang, and Chen~Change Loy.
\newblock Domain generalization: A survey.
\newblock \emph{IEEE Transactions on Pattern Analysis and Machine Intelligence}, 45\penalty0 (4):\penalty0 4396--4415, 2022{\natexlab{a}}.

\bibitem[Zhou et~al.(2016)Zhou, Wu, Ni, Zhou, Wen, and Zou]{zhou2016dorefa}
Shuchang Zhou, Yuxin Wu, Zekun Ni, Xinyu Zhou, He~Wen, and Yuheng Zou.
\newblock Dorefa-net: Training low bitwidth convolutional neural networks with low bitwidth gradients.
\newblock \emph{arXiv preprint arXiv:1606.06160}, 2016.

\bibitem[Zhou et~al.(2022{\natexlab{b}})Zhou, Wang, Konukoglu, and Van~Gool]{zhou2022rethinking}
Tianfei Zhou, Wenguan Wang, Ender Konukoglu, and Luc Van~Gool.
\newblock Rethinking semantic segmentation: A prototype view.
\newblock In \emph{Proceedings of the IEEE/CVF Conference on Computer Vision and Pattern Recognition}, pp.\  2582--2593, 2022{\natexlab{b}}.

\bibitem[Zhu et~al.(2020)Zhu, Su, Lu, Li, Wang, and Dai]{zhu2020deformable}
Xizhou Zhu, Weijie Su, Lewei Lu, Bin Li, Xiaogang Wang, and Jifeng Dai.
\newblock Deformable detr: Deformable transformers for end-to-end object detection.
\newblock \emph{arXiv preprint arXiv:2010.04159}, 2020.

\bibitem[Zhuang et~al.(2022)Zhuang, Gong, Yuan, Cui, Adam, Dvornek, s~Duncan, Liu, et~al.]{zhuang2022surrogate}
Juntang Zhuang, Boqing Gong, Liangzhe Yuan, Yin Cui, Hartwig Adam, Nicha~C Dvornek, James s~Duncan, Ting Liu, et~al.
\newblock Surrogate gap minimization improves sharpness-aware training.
\newblock In \emph{International Conference on Learning Representations}, 2022.

\end{thebibliography}
